\documentclass{article}

\usepackage{iclr2019_conference,times}
\iclrfinalcopy

\usepackage{amsmath,amsfonts,bm}

\def\eqref#1{equation~\ref{#1}}

\def\1{\bm{1}}

\DeclareMathAlphabet{\mathsfit}{\encodingdefault}{\sfdefault}{m}{sl}
\SetMathAlphabet{\mathsfit}{bold}{\encodingdefault}{\sfdefault}{bx}{n}

\usepackage[utf8]{inputenc} 
\usepackage[T1]{fontenc}    
\usepackage{hyperref}       
\usepackage{url}            
\usepackage{booktabs}       
\usepackage{amsfonts}       
\usepackage{nicefrac}       
\usepackage{microtype}      
\usepackage{placeins}
\usepackage{comment}
\usepackage{courier}

\usepackage{subcaption}
\usepackage{amsmath}
\usepackage{amssymb}
\usepackage{graphicx}

\title{Latent Convolutional Models}

\author{
{ShahRukh Athar \qquad\qquad
Evgeny Burnaev \qquad\qquad
Victor Lempitsky\thanks{Currently also with Samsung AI Center, Moscow.}}\\
Skolkovo Institute of Science and Technology (Skoltech), Russia
}

\newcommand{\fig}[1]{Figure~\ref{fig:#1}}
\newcommand{\sect}[1]{Section~\ref{sect:#1}}
\newcommand{\tab}[1]{Table~\ref{tab:#1}}

\newcommand{\eq}[1]{(\ref{eq:#1})}

\begin{document}

\maketitle \vspace{-7mm}


\begin{abstract}
We present a new latent model of natural images that can be learned on large-scale datasets. The learning process provides a latent embedding for every image in the training dataset, as well as a deep convolutional network that maps the latent space to the image space. After training, the new model provides a strong and universal image prior for a variety of image restoration tasks such as large-hole inpainting, superresolution, and colorization. To model high-resolution natural images, our approach uses latent spaces of very high dimensionality (one to two orders of magnitude higher than previous latent image models). To tackle this high dimensionality, we use latent spaces with a special manifold structure (convolutional manifolds) parameterized by a ConvNet of a certain architecture. In the experiments, we compare the learned latent models with latent models learned by autoencoders, advanced variants of generative adversarial networks, and a strong baseline system using simpler parameterization of the latent space. Our model outperforms the competing approaches over a range of restoration tasks.
\end{abstract}\vspace{-5mm}

\section{Introduction}
\label{sect:intro}

Learning good image priors is one of the core problems of computer vision and machine learning. One promising approach to obtaining such priors is to learn a deep latent model, where the set of natural images is parameterized by a certain simple-structured set or probabilistic distribution, whereas the complexity of natural images is tackled by a deep ConvNet (often called a generator or a decoder) that maps from the latent space into the space of images. The best known examples are generative adversarial networks (GANs)~\citep{GAN} and autoencoders~\citep{GoodfellowDL}. 

Given a good deep latent model, virtually any image restoration task can be solved by finding a latent representation that best corresponds to the image evidence (e.g.\ the known pixels of an occluded image or a low-resolution image). The attractiveness of such approach is in the universality of the learned image prior. Indeed, applying the model to a new restoration task can be performed by simply changing the likelihood objective. The same latent model can therefore be reused for multiple tasks, and the learning process needs not to know the image degradation process in advance. This is in contrast to task-specific approaches that usually train deep feed-forward ConvNets for individual tasks, and which have a limited ability to generalize across tasks (e.g.\ a feed-forward network trained for denoising cannot perform large-hole inpainting and vice versa).

At the moment, such image restoration approach based on latent models is limited to low-resolution images. E.g.\ \citep{GAN:inpainting} showed how a latent model trained with GAN can be used to perform inpainting of tightly-cropped $64\times{}64$ face images. Below, we show that such models trained with GANs cannot generalize to higher resolution (eventhough GAN-based systems are now able to obtain high-quality samples at high resolutions \citep{Karras18}).  We argue that it is the limited dimensionality of the latent space in GANs and other existing latent models that precludes them from spanning the space of high-resolution natural images.

To scale up latent modeling to high-resolution images, we consider latent models with tens of thousands of latent dimensions (as compared to few hundred latent dimensions in existing works). We show that training such latent models is possible using direct optimization~\citep{Bojanowski17} and that such training leads to good image priors that can be used across a broad variety of reconstruction tasks. In previous models, the latent space has a simple structure such as a sphere or a box in a Euclidean space, or a full Euclidean space with a Gaussian prior. Such choice, however, is not viable in our case, as vectors with tens of thousands of dimensions cannot be easily used as inputs to a generator. Therefore, we consider two alternative parameterizations of a latent space. Firstly, as a baseline, we consider latent spaces parameterized by image stacks (three-dimensional tensors), which allows to have ``fully-convolutional'' generators with reasonable number of parameters.

Our full system uses a more sophisticated parameterization of the latent space, which we call a \textit{convolutional manifold}, where the elements of the manifold correspond to the parameter vector of a separate ConvNet. Such indirect parameterization of images and image stacks have recently been shown to impose a certain prior~\citep{Ulyanov18}, which is beneficial for restoration of natural images. In our case, we show that a similar prior can be used with success to parameterize high-dimensional latent spaces.

To sum up, our contributions are as follows. Firstly, we consider the training of deep latent image models with the latent dimensionality that is much higher than previous works, and demonstrate that the resulting models provide universal (w.r.t.\ restoration tasks) image priors. Secondly, we suggest and investigate the convolutional parameterization for the latent spaces of such models, and show the benefits of such parameterization.

\begin{figure}
\centering
    \begin{subfigure}[b]{.15\linewidth}
    \includegraphics[width=1.03\linewidth]{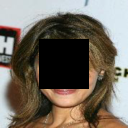}
    \end{subfigure}
    \begin{subfigure}[b]{.15\linewidth}
    \includegraphics[width=1.03\linewidth]{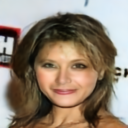}
    \end{subfigure}
    \begin{subfigure}[b]{.15\linewidth}
    \includegraphics[width=1.03\linewidth]{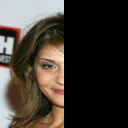}
    \end{subfigure}
    \begin{subfigure}[b]{.15\linewidth}
    \includegraphics[width=1.03\linewidth]{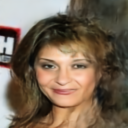}
    \end{subfigure}
    \begin{subfigure}[b]{.15\linewidth}
    \includegraphics[width=1.03\linewidth]{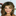}
    \end{subfigure}
    \begin{subfigure}[b]{.15\linewidth}
    \includegraphics[width=1.03\linewidth]{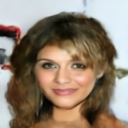}
    \end{subfigure}
    \caption{Restorations using the same Latent Convolutional Model (images 2,4,6) for different image degradations (images 1,3,5). At training time, our approach builds a latent model of undegraded images, and at test time the restoration process simply finds a latent representation that maximizes the likelihood of the corrupted image.}\vspace{-5mm}
     \label{fig:teaser}
\end{figure}

Our experiments are performed on CelebA~\citep{CelebA}~(128x128 resolution), SUN Bedrooms~\citep{lsun}~(256x256 resolution), CelebA-HQ~\citep{Karras18}~ (1024x1024 resolution) datasets, and we demonstrate that the latent models, once trained, can be applied to large hole inpainting, superresolution of very small images, and colorization tasks, outperforming other latent models in our comparisons. To the best of our knowledge, we are the first to demonstrate how ``direct'' latent modeling of natural images without extra components can be used to solve image restoration problems at these resolutions (\fig{teaser}).

\paragraph{Other related work.} Deep latent models follow a long line of works on latent image models that goes back at least to the eigenfaces approach~\citep{Sirovich87}. In terms of restoration, a competing and more popular approach are feed-forward networks trained for specific restoration tasks, which have seen rapid progress recently. Our approach does not quite match the quality of e.g.~\citep{IizukaSIGGRAPH2017}, that is designed and trained specifically for the inpainting task, or the quality of e.g.~\citep{yu2016ultra} that is designed and trained specifically for the face superresolution task. Yet the models trained within our approach (like other latent models) are universal, as they can handle degradations unanticipated at training time.

Our work is also related to pre-deep learning (``shallow'') methods that learn priors on (potentially-overlapping) image patches using maximum likelihood-type objectives such as \citep{Roth05,Karklin09,Zoran11}. The use of multiple layers in our method allows to capture much longer correlations. As a result, our method can be used successfully to handle restoration tasks that require exploiting these correlations, such as large-hole inpainting.

\section{Method}
\label{sect:method}

\newcommand{\x}{{\mathbf{x}}}
\newcommand{\z}{{\mathbf{z}}}
\newcommand{\s}{{\mathbf{s}}}
\newcommand{\y}{{\mathbf{y}}}

\newcommand{\Xd}{{X}}
\newcommand{\Xs}{{\mathcal{X}}}
\newcommand{\Zd}{{Z}}
\newcommand{\Zs}{{\mathcal{Z}}}

Let $\{\x_1,\x_2,\dots,\x_N\}$ be a set of training images, that are considered to be samples from the distribution $\Xd$ of images in the space $\Xs$ of images of a certain size that need to be modeled. In latent modeling, we introduce a different space $\Zs$ and a certain distribution $\Zd$ in that space that are used to re-parameterize $\Xs$. In previous works, $\Zs$ is usually chosen to be a Euclidean space with few dozen to few hundred dimensions, while our choice for $\Zs$ is discussed further below.

The deep latent modeling of images implies learning the generator network $g_\theta$ with learnable parameters $\theta$, which usually has convolutional architecture. The generator network maps from $\Zs$ to $\Xs$ and in particular is trained so that $g_\theta(\Zd) \approx \Xd$. Achieving the latter condition is extremely hard, and there are several approaches that can be used. Thus, generative adversarial networks (GANs)~\citep{GAN} train the generator network in parallel with a separate discriminator network that in some variants of GANs serves as an approximate ratio estimator between $\Xd$ and $\Xd +g_\theta(\Zd)$ over points in $\Xs$. Alternatively, autoencoders \citep{GoodfellowDL} and their variational counter-parts \citep{VAE} train the generator in parallel with the encoder operating in the reverse direction, resulting in a more complex distribution $\Zd$. Of these two approaches, only GANs are known to be capable of synthesizing high-resolution images, although such ability comes with additional tricks and modifications of the learning formulation \citep{WGAN,Karras18}.
In this work, we start with a simpler approach to deep latent modeling~\citep{Bojanowski17} known as the GLO model. GLO model optimizes the parameters of the generator network in parallel with the explicit embeddings of the training examples $\{\z_1,\z_2,\dots,\z_N\}$, such that $g_\theta(\z_i)\approx \x_i$ by the end of the optimization. Our approach differs from and expands \citep{Bojanowski17} in three ways: (i) we consider a much higher dimensionality of the latent space, (ii) we use an indirect parameterization of the latent space discussed further below, (iii) we demonstrate the applicability of the resulting model to a variety of image restoration tasks.

\begin{figure}[t]
    \centering
    \includegraphics[scale=0.25]{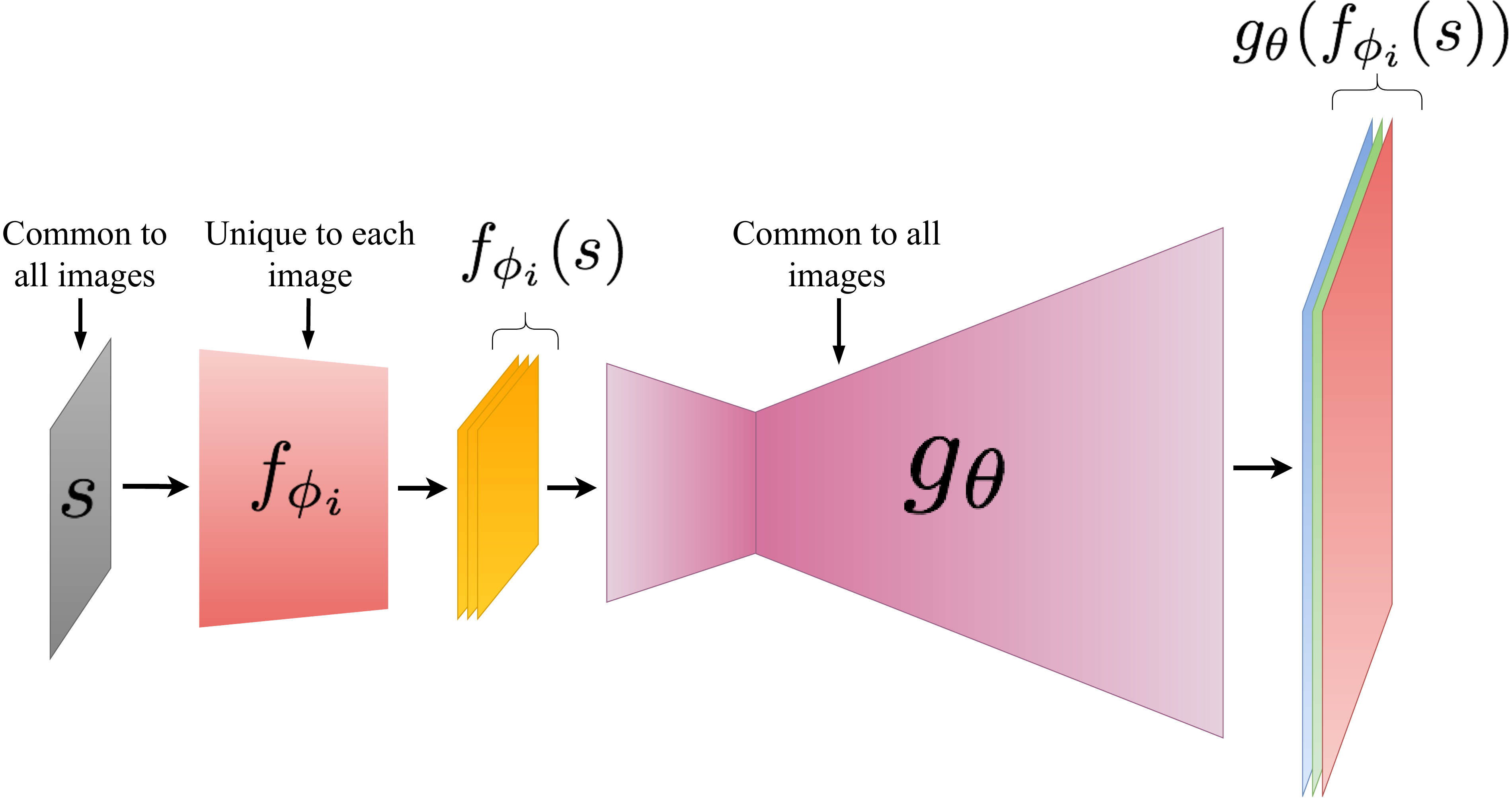}
    \caption{The Latent Convolutional Model incroprorates two sequential ConvNets. The smaller ConvNet $f$ (red) is fitted to each training image and is effectively used to parameterize the latent manifold. The bigger ConvNet $g$ (magenta) is used as a generator, and its parameters are fitted to all training data. The input $s$ to the pipeline is fixed to a random noise and not updated during training.}\vspace{-5mm}
    \label{fig:LCM}
\end{figure}

\paragraph{Scaling up latent modeling.}\label{para:scaling} Relatively low-dimensional latent models of natural images presented in previous works are capable of producing visually-compelling image samples from the distribution \citep{Karras18}, but are not actually capable of matching or covering a rather high-dimensional distribution $\Xd$. E.g.\ in our experiments, none of GAN models were capable of reconstructing most samples $\x$ from the hold-out set (or even from the training set; this observation is consistent with \citep{Bojanowski17} and also with \citep{EforsBags}). Being unable to reconstruct uncorrupted samples clearly suggests that the learned models are not suitable to perform restoration of corrupted samples. On the other hand, autoencoders and the related GLO latent model~\citep{Bojanowski17} were able to achieve better reconstructions than GAN on the hold-out sets, yet have distinctly blurry reconstructions (even on the training set), suggesting strong underfitting.

We posit that existing deep latent models are limited by the dimensionality of the latent space that they consider, and aim to scale up this dimensionality significantly.
Simply scaling up the latent dimensionality to few tens of dimensions is not easily feasible, as e.g.\ the generator network has to work with such a vector as an input, which would make the first fully-connected layer excessively large with hundreds of millions of parameters\footnote{One can consider the first layer having a very thin matrix with a reasonable number of parameters mapping the latent vector to a much lower-dimensional space. This however would effectively amount to using lower-dimensional latent space and would defy the idea of scaling up latent dimensionality.}.

To achieve a tractable size of the generator, one can consider latent elements $\z$ to have a three-dimensional tensor structure, i.e.\ to be stacks of 2D image maps. Such choice of structure is very natural for convolutional architectures, and allows to train ``fully-convolutional'' generators with the first layer being a standard convolutional operation. The downside of this choice, as we shall see, is that it allows limited coordination between distant parts of the images $\x=g_\theta(\z)$ produced by the generator. This drawback is avoided when the latent space is parameterized using latent convolutional manifolds as described next.

\paragraph{Latent convolutional manifolds.} To impose more appropriate structure on the latent space, we consider structuring these spaces as \textit{convolutional manifolds} defined as follows. Let $\s$ be a stack of maps of the size $W_s\times H_s \times C_s$ and let $\{f_\phi\,|\,\phi \in \Phi\}$ be a set of convolutional networks all sharing the same architecture $f$ that maps $\s$ to different maps of size $W_z\times H_z \times C_z$. A certain parameter vector $\phi \in \Phi$ thus defines a certain convolutional network $f_\phi$. Then, let $\z(\phi) = f_\phi(\s)$ be an element in the space of $(W_z\times H_z \times C_z)$-dimensional maps. Various choices of $\phi$ then span a manifold embedded into this space, and we refer to it as the \textit{convolutional manifold}. A convolutional manifold $\mathbf{C}_{f,\s}$ is thus defined by the ConvNet architecture $f$ as well as by the choice of the input $\s$ (which in our experiments is always chosen to be filled with uniform random noise). Additionally, we also restrict the elements of vectors $\phi$ to lie within the $[-B;B]$ range. Formally, the convolutional manifold is defined as the following set:
\begin{equation}
\mathbf{C}_{f,\s} = \{\z\,|\, \z = f_\phi(\s), \phi \in \Phi\}\,,\; \Phi = [-B;B]^{N_\phi}\,,
\end{equation}
where $\phi$ serves as a natural parameterization and $N_\phi$ is the number of network parameters. Below, we refer to $f$ as \textit{latent ConvNet}, to disambiguate it from the generator $g$, which also has a convolutional structure.

The idea of the convolutional manifold is inspired by the recent work on deep image priors~\citep{Ulyanov18}. While they effectively use convolutional manifolds to model natural images directly, in our case, we use them to model the latent space of the generator networks resulting in a fully-fledged learnable latent image model (whereas the model in~\citep{Ulyanov18} cannot be learned on a dataset of images). The work~\citep{Ulyanov18} demonstrates that the regularization imposed by the structure of a very high-dimensional convolutional manifold is beneficial when modeling natural images. Our intuition here is that similar regularization would be beneficial in regularizing learning of high-dimensional latent spaces. As our experiments below reveal, this intuition holds true.

\paragraph{Learning formulation.} Learning the deep latent model (\fig{LCM}) in our framework then amounts to the following optimization task. Given the training examples $\{\x_1,\x_2,\dots,\x_N\}$, the architecture $f$ of the convolutional manifold, and the architecture $g$ of the generator network, we seek the set of the latent ConvNet parameter vectors $\{\phi_1,\phi_2,\dots,\phi_N\}$ and the parameters of the generator network $\theta$ that minimize the following objective:
\begin{equation}\label{eq:obj}
L(\phi_1,\phi_2,\dots,\phi_N,\theta)=\frac{1}{N}\sum_{i=1}^N \|\, g_\theta(\,f_{\phi_i}(\s)\,) - \x_i\, \|\,,
\end{equation}
with an additional box constraints $\phi_i^j \in [-0.01;0.01]$ and $\s$ being a random set of image maps filled with uniform noise. Following~\citep{Bojanowski17}, the norm in \eq{obj} is taken to be the Laplacian-L1: $\|\x_{1} - \x_{2}\|_\text{Lap-L1} = \sum_{j}2^{-2j}|L^{j}(\x_{1} - \x_{2})|_{1}$, where $L^j$ is the $j$th level of the Laplacian image pyramid~\citep{Burt83}. We have also found that adding an extra MSE loss term to the Lap-L1 loss term with the weight of 1.0 speeds up convergence of the models without affecting the results by much.

The optimization \eq{obj} is performed using stochastic gradient descent. As an outcome of the optimization, each training example $\x_i$ gets a representation $\z_i=f_{\phi_i}$ on the convolutional manifold $\mathbf{C}_{f,\s}$.

Importantly, the elements of the convolutional manifold then define a set of images in the image space (which is the image of the convolutional manifold under learned generator):
\begin{equation}\label{eq:image_manifold}
\mathbf{I}_{f,\s,\theta} = \{\x\,|\, \x = g_\theta( f_\phi(\s)), \phi \in \Phi\}\,.
\end{equation}
\begin{figure}[ht]

    \centering
    \begin{tabular}{cc}
        \includegraphics[trim=20 100 20 60,clip,width=0.5\textwidth]{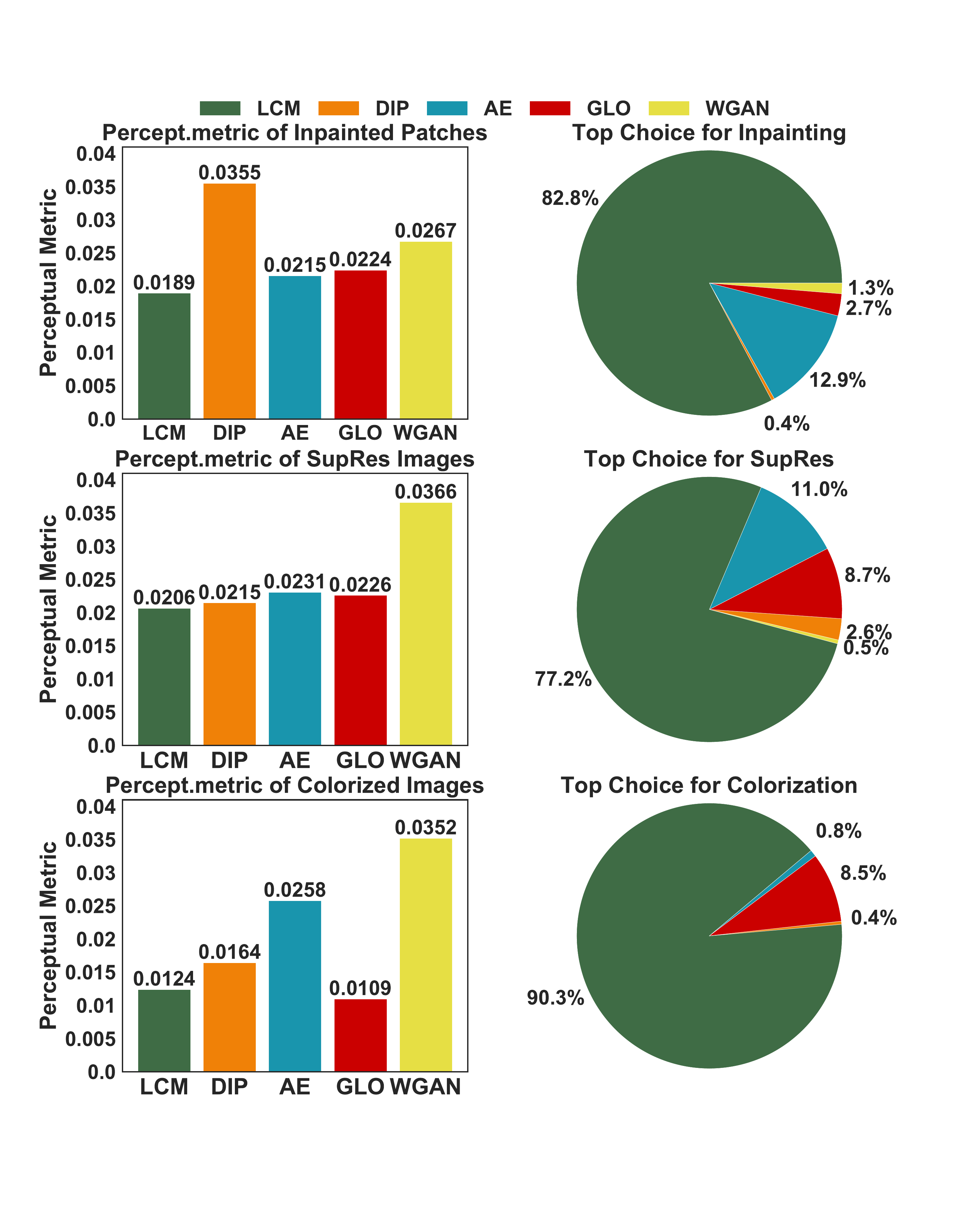}&       
        \includegraphics[trim=20 100 20 60,clip,width=0.5\textwidth]{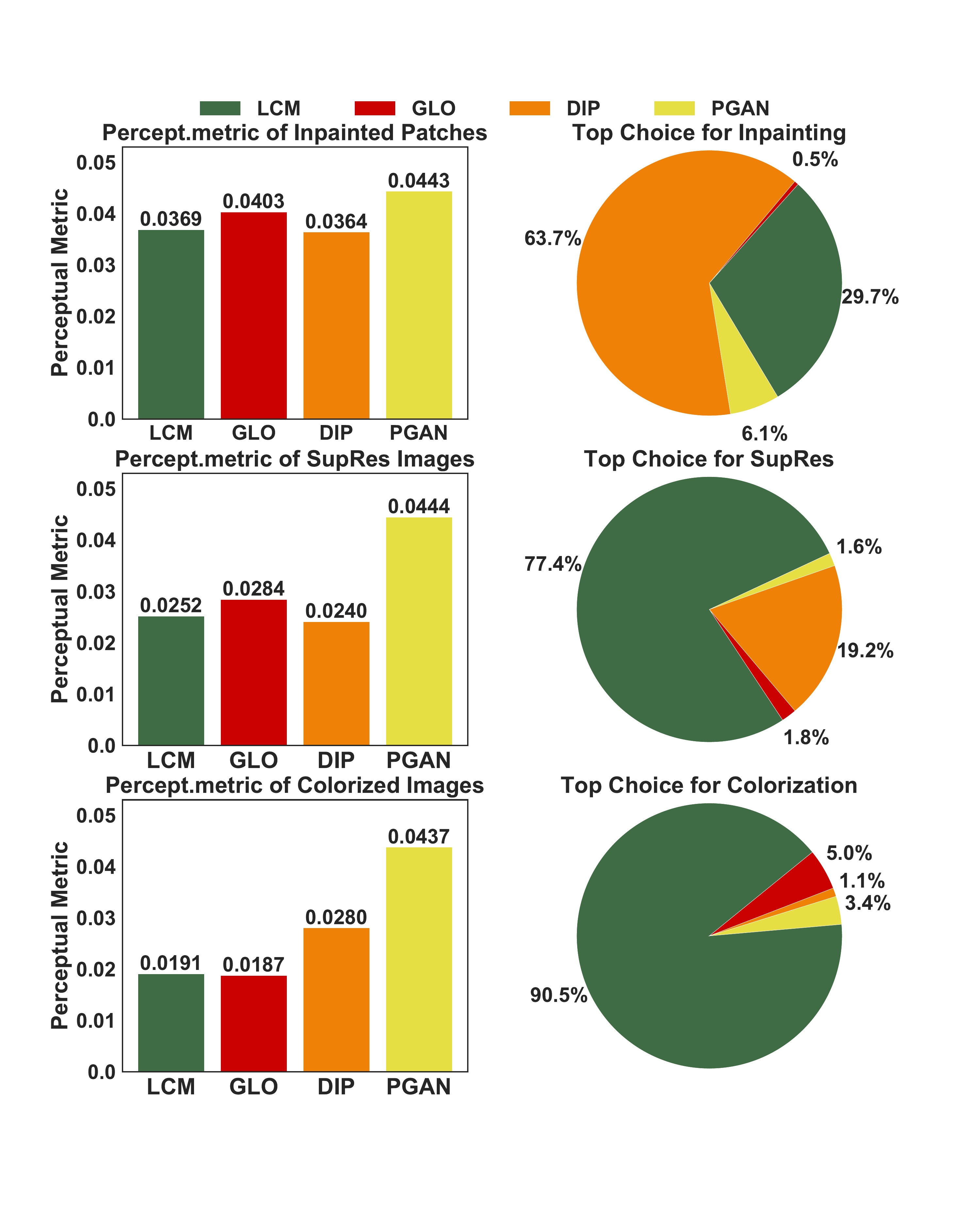}
    \end{tabular}
    \caption{Results (perceptual metrics -- lower is better -- and user preferences) for the two datasets (CelebA -- left, Bedrooms -- right) and three tasks (inpainting, super-resolution, colorization). For the colorization task the perceptual metric is inadequate as the grayscale image has the lowest error, but is shown for completeness.}\vspace{-5mm}
    \label{fig:results}
\end{figure}
\vspace{6cm}
\begin{table}[ht]
\caption{MSE loss on the restored images with respect to the ground truth. For inpainting the MSE was calculated just over the inpainted region of the images.}
\label{tab:ReconsMSE}
\resizebox{\linewidth}{!}{%
\begin{tabular}{@{}llllll|llll@{}}
\toprule
\multicolumn{6}{c}{\textbf{CelebA}}                              & \multicolumn{4}{c}{\textbf{LSUN-Bedrooms}} \\ \midrule
             & \textbf{LCM}    & \textbf{GLO}    & \textbf{DIP}    & \textbf{AE}     & \textbf{WGAN}   & \textbf{LCM}     & \textbf{GLO}    & \textbf{DIP}    & \textbf{PGAN}   \\
\textbf{Inpainting}   & 0.0034 & 0.0038 & 0.0091 & 0.0065 & 0.0344 & 0.0065  & 0.0085 & 0.0063 & 0.0097 \\
\textbf{Super-res}    & 0.0061 & 0.0063 & 0.0052 & 0.0083 & 0.0446 & 0.0071  & 0.0069 & 0.0057 & 0.0183 \\ 
\textbf{Colorization} & 0.0071 & 0.0069 & 0.0136 & 0.0194 & 0.0373 & 0.0066  & 0.0075 & 0.0696 & 0.0205 \\ \bottomrule
\end{tabular}}

\end{table}

While not all elements of the manifold $\mathbf{I}_{f,\s,\theta}$ will correspond to natural images from the distribution $\Xd$, we have found out that with few thousand dimensions, the resulting manifolds can cover the support of $\Xd$ rather well. I.e.\ each sample from the image distribution can be approximated by the element of $\mathbf{I}_{f,\s,\theta}$ with a low approximation error. This property can be used to perform all kinds of image restoration tasks.

\paragraph{Image restoration using learned latent models.} We now describe how the learned latent model can be used to perform the restoration of the unknown image $\x_0$ from the distribution $\Xd$, given some evidence $\y$. Depending on the degradation process, the evidence $\y$ can be an image $\x_0$ with masked values (inpainting task), the low-resolution version of $\x_0$ (superresolution task), the grayscale version of $\x_0$ (colorization task), the noisy version of $\x_0$ (denoising task), a certain statistics of $\x_0$ computed e.g.\ using a deep network (feature inversion task), etc.

We further assume, that the degradation process is described by the objective $E(\x|\y)$, which can be set to minus log-likelihood $E(\x|\y) = -\log p(\y|\x)$ of observing $\y$ as a result of the degradation of $\x$. E.g.\ for the \textit{inpainting} task, one can use $E(\x|\y) = \|(\x-\y)\odot \mathbf{m}\|$, where $\mathbf{m}$ is the 0-1 mask of known pixels and $\odot$ denotes element-wise product. For the \textit{superresolution} task, the restoration objective is naturally defined as $E(\x|\y) = \|\downarrow(\x)-\y\|$, where $\downarrow(\cdot)$ is an image downsampling operator (we use Lanczos in the experiments) and $\y$ is the low-resolution version of the image. For the \textit{colorization} task, the objective is defined as $E(\x|\y) = \|\text{gray}(\x)-\y\|$, where $\text{gray}(\cdot)$ denotes a projection from the RGB to grayscale images (we use a simple averaging of the three color channels in the experiments) and $\y$ is the grayscale version of the image.

Using the learned latent model as a prior, the following estimation combining the learned prior and the provided image evidence is performed:
\begin{align}\label{eq:reconstr}
\hat \phi = \arg\min_\phi E(g_\theta( f_\phi(\s))\,|\,\y)\,,\qquad
\hat \x = g_\theta(f_{\hat \phi}(\s))\,.
\end{align}
In other words, we simply estimate the element of the image  manifold \eq{image_manifold} that has the highest likelihood. The optimization is performed using stochastic gradient descent over the parameters $\phi$ on the latent convolutional manifold. 

For the baseline models, which use a direct parameterization of the latent space, we perform analogous estimation using optimization in the latent space:
\begin{align}\label{eq:reconstr_baseline}
\hat \z = \arg\min_\z E(g_\theta( \z)\,|\,\y)\,,\qquad
\hat \x = g_\theta(\z)\,.
\end{align}
In the experiments, we compare the performance of our full model and several baseline models over a range of the restoration tasks using formulations \eq{reconstr} and \eq{reconstr_baseline}.

\section{Experiments}
\label{sect:experiments}
\begin{figure}
\centering
\begin{subfigure}[b]{.13\linewidth}
\caption*{\centering\small{Distorted Image}}
\includegraphics[width=1.03\linewidth]{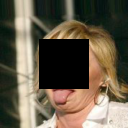}
\end{subfigure}
\begin{subfigure}[b]{.13\linewidth}
\caption*{\centering\small{LCM (Ours)}}
\includegraphics[width=1.03\linewidth]{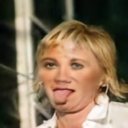}
\end{subfigure}
\begin{subfigure}[b]{.13\linewidth}
\caption*{\centering\small{GLO}}
\includegraphics[width=1.03\linewidth]{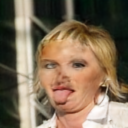}
\end{subfigure}
\begin{subfigure}[b]{.13\linewidth}
\caption*{\centering\small{DIP}}
\includegraphics[width=1.03\linewidth]{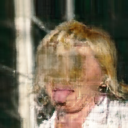}
\end{subfigure}
\begin{subfigure}[b]{.13\linewidth}
\caption*{\centering\small{WGAN}}
\includegraphics[width=1.03\linewidth]{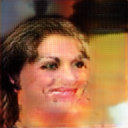}
\end{subfigure}
\begin{subfigure}[b]{.13\linewidth}
\caption*{\centering\small{AE}}
\includegraphics[width=1.03\linewidth]{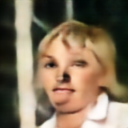}
\end{subfigure}
\begin{subfigure}[b]{.13\linewidth}
\caption*{\centering\small{Original Image}}
\includegraphics[width=1.03\linewidth]{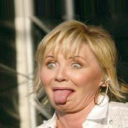}
\end{subfigure}

\begin{subfigure}[b]{.13\linewidth}
\includegraphics[width=1.03\linewidth]{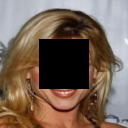}
\end{subfigure}
\begin{subfigure}[b]{.13\linewidth}
\includegraphics[width=1.03\linewidth]{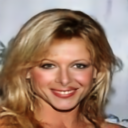}
\end{subfigure}
\begin{subfigure}[b]{.13\linewidth}
\includegraphics[width=1.03\linewidth]{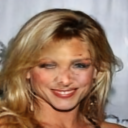}
\end{subfigure}
\begin{subfigure}[b]{.13\linewidth}
\includegraphics[width=1.03\linewidth]{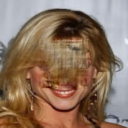}
\end{subfigure}
\begin{subfigure}[b]{.13\linewidth}
\includegraphics[width=1.03\linewidth]{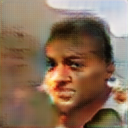}
\end{subfigure}
\begin{subfigure}[b]{.13\linewidth}
\includegraphics[width=1.03\linewidth]{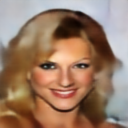}
\end{subfigure}
\begin{subfigure}[b]{.13\linewidth}
\includegraphics[width=1.03\linewidth]{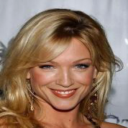}
\end{subfigure}


\begin{subfigure}[b]{.13\linewidth}
\includegraphics[width=1.03\linewidth]{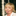}
\end{subfigure}
\begin{subfigure}[b]{.13\linewidth}
\includegraphics[width=1.03\linewidth]{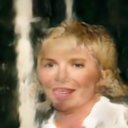}
\end{subfigure}
\begin{subfigure}[b]{.13\linewidth}
\includegraphics[width=1.03\linewidth]{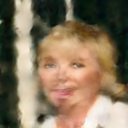}
\end{subfigure}
\begin{subfigure}[b]{.13\linewidth}
\includegraphics[width=1.03\linewidth]{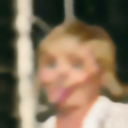}
\end{subfigure}
\begin{subfigure}[b]{.13\linewidth}
\includegraphics[width=1.03\linewidth]{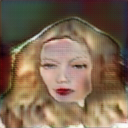}
\end{subfigure}
\begin{subfigure}[b]{.13\linewidth}
\includegraphics[width=1.03\linewidth]{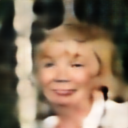}
\end{subfigure}
\begin{subfigure}[b]{.13\linewidth}
\includegraphics[width=1.03\linewidth]{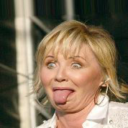}
\end{subfigure}

\begin{subfigure}[b]{.13\linewidth}
\includegraphics[width=1.03\linewidth]{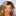}
\end{subfigure}
\begin{subfigure}[b]{.13\linewidth}
\includegraphics[width=1.03\linewidth]{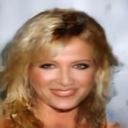}
\end{subfigure}
\begin{subfigure}[b]{.13\linewidth}
\includegraphics[width=1.03\linewidth]{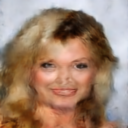}
\end{subfigure}
\begin{subfigure}[b]{.13\linewidth}
\includegraphics[width=1.03\linewidth]{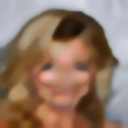}
\end{subfigure}
\begin{subfigure}[b]{.13\linewidth}
\includegraphics[width=1.03\linewidth]{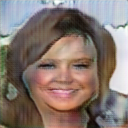}
\end{subfigure}
\begin{subfigure}[b]{.13\linewidth}
\includegraphics[width=1.03\linewidth]{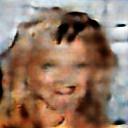}
\end{subfigure}
\begin{subfigure}[b]{.13\linewidth}
\includegraphics[width=1.03\linewidth]{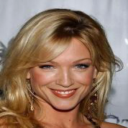}
\end{subfigure}




\begin{subfigure}[b]{.13\linewidth}
\includegraphics[width=1.03\linewidth]{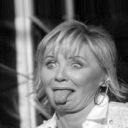}
\end{subfigure}
\begin{subfigure}[b]{.13\linewidth}
\includegraphics[width=1.03\linewidth]{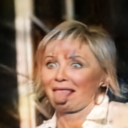}
\end{subfigure}
\begin{subfigure}[b]{.13\linewidth}
\includegraphics[width=1.03\linewidth]{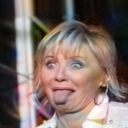}
\end{subfigure}
\begin{subfigure}[b]{.13\linewidth}
\includegraphics[width=1.03\linewidth]{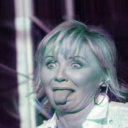}
\end{subfigure}
\begin{subfigure}[b]{.13\linewidth}
\includegraphics[width=1.03\linewidth]{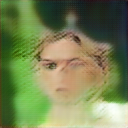}
\end{subfigure}
\begin{subfigure}[b]{.13\linewidth}
\includegraphics[width=1.03\linewidth]{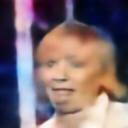}
\end{subfigure}
\begin{subfigure}[b]{.13\linewidth}
\includegraphics[width=1.03\linewidth]{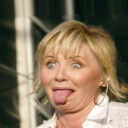}
\end{subfigure}

\begin{subfigure}[b]{.13\linewidth}
\includegraphics[width=1.03\linewidth]{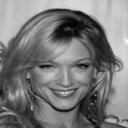}
\end{subfigure}
\begin{subfigure}[b]{.13\linewidth}
\includegraphics[width=1.03\linewidth]{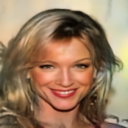}
\end{subfigure}
\begin{subfigure}[b]{.13\linewidth}
\includegraphics[width=1.03\linewidth]{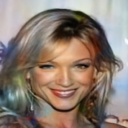}
\end{subfigure}
\begin{subfigure}[b]{.13\linewidth}
\includegraphics[width=1.03\linewidth]{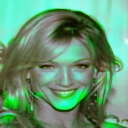}
\end{subfigure}
\begin{subfigure}[b]{.13\linewidth}
\includegraphics[width=1.03\linewidth]{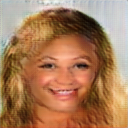}
\end{subfigure}
\begin{subfigure}[b]{.13\linewidth}
\includegraphics[width=1.03\linewidth]{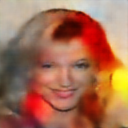}
\end{subfigure}
\begin{subfigure}[b]{.13\linewidth}
\includegraphics[width=1.03\linewidth]{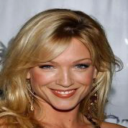}
\end{subfigure}


\caption{Qualitative comparison on CelebA (see the text for discussion).}\vspace{-5mm}

\label{fig:celeba}
\end{figure}

\begin{figure}[ht]
\centering

\begin{subfigure}[b]{.15\linewidth}
\caption*{\centering\small{Distorted Image}}
\includegraphics[width=1.03\linewidth]{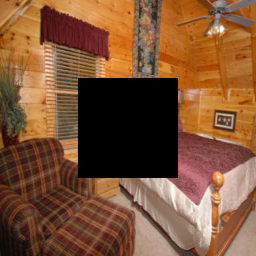}
\end{subfigure}
\begin{subfigure}[b]{.15\linewidth}
\caption*{\centering\small{LCM (Ours)}}
\includegraphics[width=1.03\linewidth]{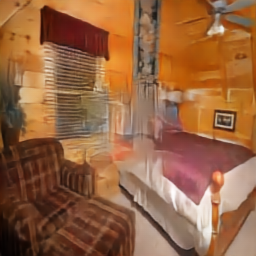}
\end{subfigure}
\begin{subfigure}[b]{.15\linewidth}
\caption*{\centering\small{GLO}}
\includegraphics[width=1.03\linewidth]{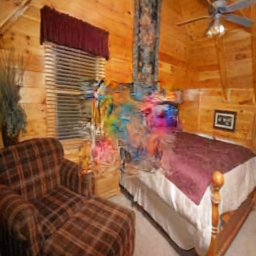}
\end{subfigure}
\begin{subfigure}[b]{.15\linewidth}
\caption*{\centering\small{DIP}}
\includegraphics[width=1.03\linewidth]{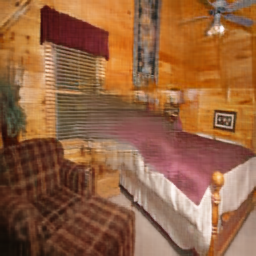}
\end{subfigure}
\begin{subfigure}[b]{.15\linewidth}
\caption*{\centering\small{PGAN}}
\includegraphics[width=1.03\linewidth]{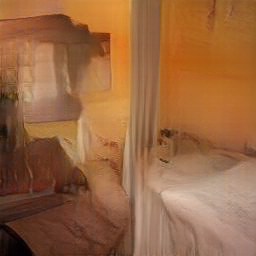}
\end{subfigure}
\begin{subfigure}[b]{.15\linewidth}
\caption*{\centering\small{Original Image}}
\includegraphics[width=1.03\linewidth]{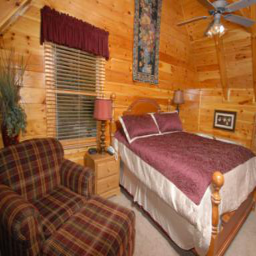}
\end{subfigure}

\begin{subfigure}[b]{.15\linewidth}
\includegraphics[width=1.03\linewidth]{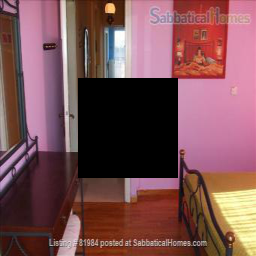}
\end{subfigure}
\begin{subfigure}[b]{.15\linewidth}
\includegraphics[width=1.03\linewidth]{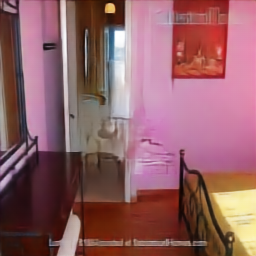}
\end{subfigure}
\begin{subfigure}[b]{.15\linewidth}
\includegraphics[width=1.03\linewidth]{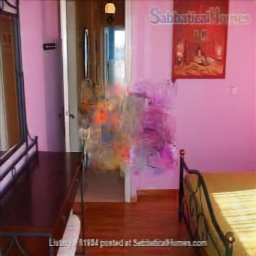}
\end{subfigure}
\begin{subfigure}[b]{.15\linewidth}
\includegraphics[width=1.03\linewidth]{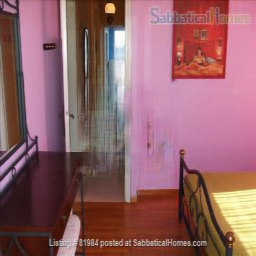}
\end{subfigure}
\begin{subfigure}[b]{.15\linewidth}
\includegraphics[width=1.03\linewidth]{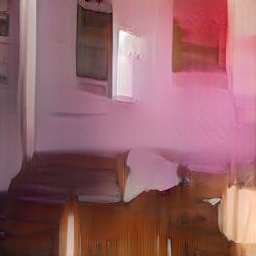}
\end{subfigure}
\begin{subfigure}[b]{.15\linewidth}
\includegraphics[width=1.03\linewidth]{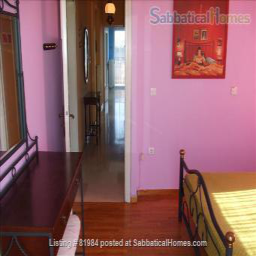}
\end{subfigure}

\begin{subfigure}[b]{.15\linewidth}
\includegraphics[width=1.03\linewidth]{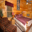}
\end{subfigure}
\begin{subfigure}[b]{.15\linewidth}
\includegraphics[width=1.03\linewidth]{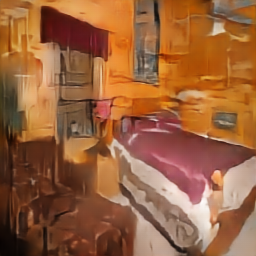}
\end{subfigure}
\begin{subfigure}[b]{.15\linewidth}
\includegraphics[width=1.03\linewidth]{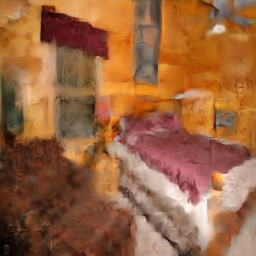}
\end{subfigure}
\begin{subfigure}[b]{.15\linewidth}
\includegraphics[width=1.03\linewidth]{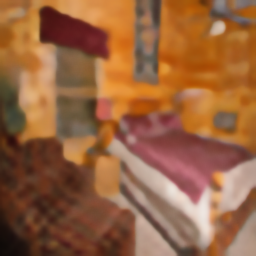}
\end{subfigure}
\begin{subfigure}[b]{.15\linewidth}
\includegraphics[width=1.03\linewidth]{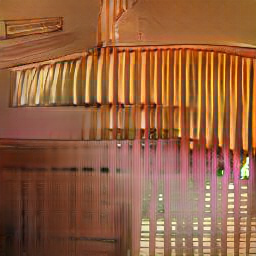}
\end{subfigure}
\begin{subfigure}[b]{.15\linewidth}
\includegraphics[width=1.03\linewidth]{figures/Bedrooms/Org_Images/Org_image5.png}
\end{subfigure}

\begin{subfigure}[b]{.15\linewidth}
\includegraphics[width=1.03\linewidth]{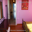}
\end{subfigure}
\begin{subfigure}[b]{.15\linewidth}
\includegraphics[width=1.03\linewidth]{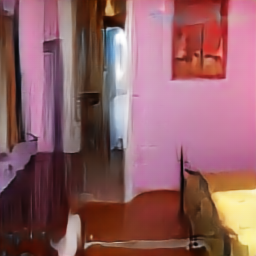}
\end{subfigure}
\begin{subfigure}[b]{.15\linewidth}
\includegraphics[width=1.03\linewidth]{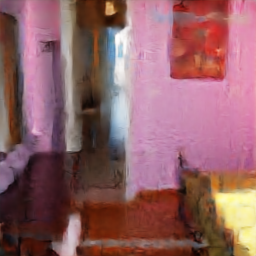}
\end{subfigure}
\begin{subfigure}[b]{.15\linewidth}
\includegraphics[width=1.03\linewidth]{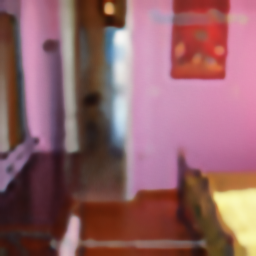}
\end{subfigure}
\begin{subfigure}[b]{.15\linewidth}
\includegraphics[width=1.03\linewidth]{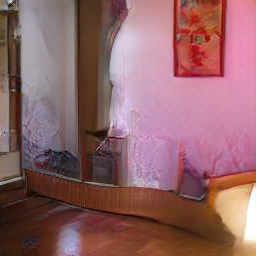}
\end{subfigure}
\begin{subfigure}[b]{.15\linewidth}
\includegraphics[width=1.03\linewidth]{figures/Bedrooms/Org_Images/Org_image13.png}
\end{subfigure}

\begin{subfigure}[b]{.15\linewidth}
\includegraphics[width=1.03\linewidth]{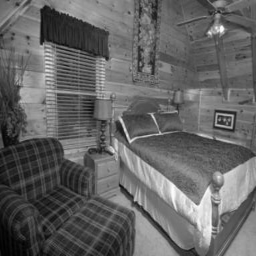}
\end{subfigure}
\begin{subfigure}[b]{.15\linewidth}
\includegraphics[width=1.03\linewidth]{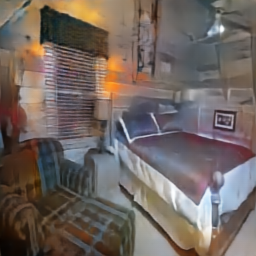}
\end{subfigure}
\begin{subfigure}[b]{.15\linewidth}
\includegraphics[width=1.03\linewidth]{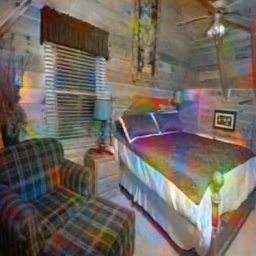}
\end{subfigure}
\begin{subfigure}[b]{.15\linewidth}
\includegraphics[width=1.03\linewidth]{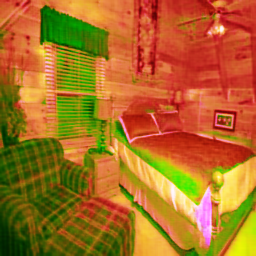}
\end{subfigure}
\begin{subfigure}[b]{.15\linewidth}
\includegraphics[width=1.03\linewidth]{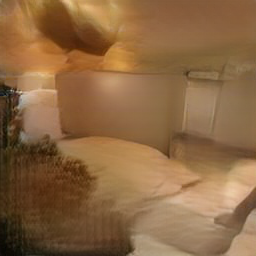}
\end{subfigure}
\begin{subfigure}[b]{.15\linewidth}
\includegraphics[width=1.03\linewidth]{figures/Bedrooms/Org_Images/Org_image5.png}
\end{subfigure}

\begin{subfigure}[b]{.15\linewidth}
\includegraphics[width=1.03\linewidth]{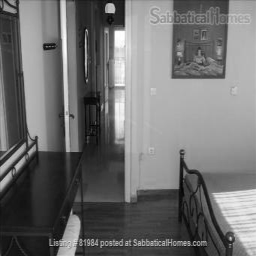}
\end{subfigure}
\begin{subfigure}[b]{.15\linewidth}
\includegraphics[width=1.03\linewidth]{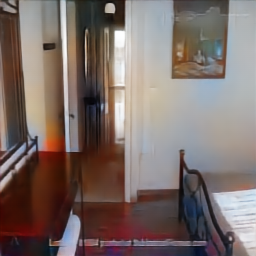}
\end{subfigure}
\begin{subfigure}[b]{.15\linewidth}
\includegraphics[width=1.03\linewidth]{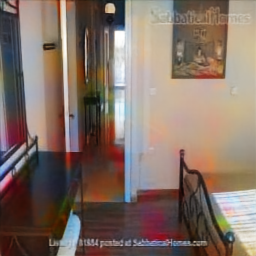}
\end{subfigure}
\begin{subfigure}[b]{.15\linewidth}
\includegraphics[width=1.03\linewidth]{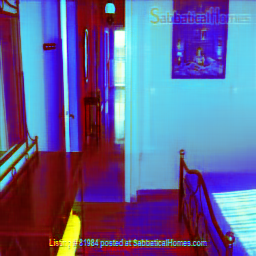}
\end{subfigure}
\begin{subfigure}[b]{.15\linewidth}
\includegraphics[width=1.03\linewidth]{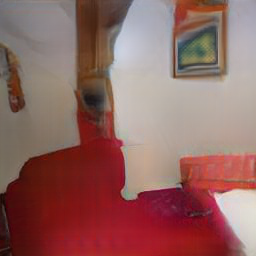}
\end{subfigure}
\begin{subfigure}[b]{.15\linewidth}
\includegraphics[width=1.03\linewidth]{figures/Bedrooms/Org_Images/Org_image13.png}
\end{subfigure}

\caption{Qualitative comparison on SUN Bedrooms for the tasks of inpainting (rows 1-2), superresolution (rows 3-4), colorization (rows 5-6). The LCM method performs better than most methods for the first two tasks.}

\label{fig:bedrooms}
\end{figure}

\paragraph{Datasets.} The experiments were conducted on three datasets. The \textbf{CelebA} dataset was obtained by taking the 150K images from \citep{CelebA} (cropped version) and resizing them from 178$\times$218 to $128\times{}128$. Note that unlike most other works, we have performed anisotropic rescaling rather than additional cropping, leading to the version of the dataset with larger background portions and higher variability (corresponding to a harder modeling task). The \textbf{Bedrooms} dataset from the LSUN~\citep{lsun} is another popular dataset of images. We rescale all images to the $256\times{}256$ size. Finally, the \textbf{CelebA-HQ} dataset from \citep{Karras18} that consists of 30K $1024\times{}1024$ images of faces. 

\paragraph{Tasks.} We have compared methods for three diverse tasks. For the inpainting task, we have degraded the input images by masking the center part of the image ($50\times{}50$ for CelebA, $100\times{}100$ for Bedrooms, $400\times{}400$ for CelebA-HQ). For the superresolution task, we downsampled the images by a factor of eight. For the colorization task, we have averaged the color channels obtaining the gray version of the image.

\subsection{Experiments on CelebA and Bedrooms}
We have performed extensive comparisons with other latent models on the two datasets with smaller image size and lower training times (CelebA and Bedrooms).
The following latent models were compared:
\begin{itemize}
    \item \textbf{Latent Convolutional Networks (LCM -- Ours):} Each $f_{\phi_{i}}$ has 4 layers (in CelebA) or 5 layers (in Bedrooms) or 7 layers (in CelebA-HQ) and takes as input random uniform noise. The Generator, $g_\theta$ has an hourglass architecture. The latent dimensionality of the model was 24k for CelebA and 61k for Bedrooms.
    \item \textbf{GLO}:  The baseline model discussed in the end of \sect{method} and inspired by \citep{Bojanowski17}, where the generator network has the same architecture as in LCM, but the convolutional space is parameterized by a set of maps. The latent dimensionality is the same as in LCM (and thus much higher than in \citep{Bojanowski17}). We have also tried a variant reproduced exactly from \citep{Bojanowski17} with vectorial latent spaces that feed into a fully-connected layers (for the dimensionalities ranging from 2048 to 8162 -- see Appendix~\ref{app:glo}), but invariably observed underfitting. Generally, we took extra care to find the optimal parameterization that would be most favourable to this baseline. 
    \item \textbf{DIP}: The deep image prior-based restoration~\citep{Ulyanov18}. We use the architecture proposed by the authors in the paper. DIP can be regarded as an extreme version of our paper with the generator network being an identity. DIP fits 1M parameters to each image for inpainting and colorization and 2M parameters for super-resolution. 
    \item \textbf{GAN}: For CelebA we train a WGAN-GP \citep{WGAN-GP} with the DCGAN type generator and a latent space of 256. For Bedrooms we use the pretrained Progressive GAN (PGAN) models with the latent space of dimensionality $512$ published by the authors of \citep{Karras18}. During restoration, we do not impose prior on the norm of $\textbf{z}$ since it worsens the underfitting problem of GANs (as demonstrated in \href{https://drive.google.com/file/d/1EAnnAiUyJkIpCKtLhaG4Of7xgvLj58oZ/view}{Appendix C}).
    \item \textbf{AE}: For the CelebA we have also included a standard autoencoder using the Lap-L1 and MSE reconstruction metrics into the comparison (latent dimensionality 1024). We have also tried the variant with convolutional higher-dimensional latent space, but have observed very strong overfitting. The variational variant (latent dimensionality 1024) lead to stronger underfitting than the non-variational variant. As the experiments on CelebA clearly showed a strong underfitting, we have not included AE into the comparison on the higher-resolution Bedrooms dataset.
\end{itemize}
For Bedrooms dataset we restricted training to the first 200K training samples, except for the DIP (which does not require training) and GAN (we used the progressive GAN model trained on all 3M samples). All comparisons were performed on hold-out sets not used for training. Following \citep{Bojanowski17}, we use plain SGD with very high learning rate of 1.0 to train LCM and of 10.0 to train the GLO models. The exact architectures are given in \href{https://drive.google.com/file/d/1EAnnAiUyJkIpCKtLhaG4Of7xgvLj58oZ/view}{Appendix D}.

\paragraph{Metrics.} We have used quantitative and user study-based assessment of the results. For the quantitative measure, we have chosen the mean squared error (MSE) measure in pixel space, as well as the mean squared distance of the VGG16-features~\citep{Simonyan14} between the original and the reconstructed images. Such \textit{perceptual metrics} are known to be correlated with the human judgement \citep{Johnson16,Zhang18}. We have used the [\texttt{relu1\_2, relu2\_2, relu3\_3, relu4\_3, relu5\_3}] layers contributing to the distance metric with equal weight. Generally, we observed that the relative performance of the methods were very similar for the MSE measure, for the individual VGG layers, and for the averaged VGG metrics that we report here. When computing the loss for the inpainting task we only considered the positions corresponding to the masked part.

Quantitative metrics however have limited relevance for the tasks with big multimodal conditional distributions, i.e.\ where two very different answers can be equally plausible, such as all three tasks that we consider (e.g.\ there could be very different colorizations of the same bedroom image).

In this situation, human judgement of quality is perhaps the best measure of the algorithm performance. To obtain such judgements, we have performed a user study, where we have picked 10 random images for each of the two datasets and each of the three tasks. The results of all compared methods alongside the degraded inputs were shown to the participants (100 for CelebA, 38 for Bedrooms). For each example, each subject was asked to pick the best restoration variant (we asked to take into account both realism and fidelity to the input). The results were presented in random order (shuffled independently for each example). We then just report the percentage of user choices for each method for a given task on a given dataset averaged over all subjects and all ten images.

\paragraph{Results.} The results of the comparison are summarized in \fig{results} and \tab{ReconsMSE} with representative examples shown in \fig{celeba} and \fig{bedrooms}. ``Traditional'' latent models (built WGAN/PGAN and AE) performed poorly. In particular, GAN-based models produced results that were both unrealistic and poorly fit the likelihood. Note that during fitting we have not imposed the Gaussian prior on the latent space of GANs. Adding such prior did not result in considerable increase of realism and lead to even poorer fit to the evidence (see \href{https://drive.google.com/file/d/1EAnnAiUyJkIpCKtLhaG4Of7xgvLj58oZ/view}{Appendix C}). 

The DIP model did very well for inpainting and superresolution of relatively unstructured Bedrooms dataset. It however performed very poorly on CelebA due to its inability to learn face structure from data and on the colorization task due to its inability to learn about natural image colors.

Except for the Bedrooms-inpainting, the new models with very large latent space produced results that were clearly favoured by the users. LCM performed better than GLO in all six user comparisons, while in terms of the perceptual metric the comparison the performance of LCM was also better than GLO for inpainting and superresolution tasks. For the colorization task, the LCM is unequivocally better in terms of user preferences, and worse in terms of the perceptual metric. We note that, however, perceptual metric is inadequate for the colorization task as the original grayscale image scores better than the results of all evaluated methods. We therefore only provide the results in this metric for colorization for the sake of completeness (finding good quantitative measure for the highly-ambiguous colorization task is a well-known unsolved problem).

Additional results on CelebA and Bedrooms dataset are given in \href{https://drive.google.com/file/d/1EAnnAiUyJkIpCKtLhaG4Of7xgvLj58oZ/view}{Appendices A,F}.

\begin{table}[h!]
\caption{Metrics of optimization over the z-space, the convolutional manifold and Progressive GAN \citep{Karras18} latent space}
\label{tab:ConvMan}
\resizebox{\linewidth}{!}{%
\begin{tabular}{@{}llll@{}}
\toprule
\textbf{Optimization Over}           & \textbf{MSE (known pixels)} & \textbf{MSE (inpainted pixels)} & \textbf{Perceptual Metric} \\ \midrule
\textbf{Convolutional Net Parameters} & 0.00307               & 0.00171                   & 0.02381           \\
\textbf{Z-Space }                     & 0.00141               & 0.00854                   & 0.02736           \\
\textbf{PGAN Latent Space}            & 0.00477               & 0.00224                   & 0.02546           \\ \bottomrule
\end{tabular}}

\end{table}

\begin{figure}[h]
    \centering
    \renewcommand{\tabcolsep}{0.5pt}
    \begin{tabular}{ccccc}
    \small{Distorted Image} & \small{OptConv} & \small{OptZ} & \small{PGAN} & \small{Original Image}\\
    \includegraphics[width=0.2\linewidth]{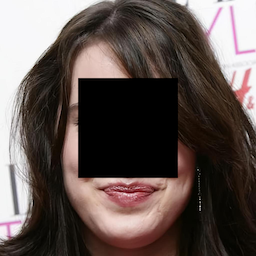}&
    \includegraphics[width=0.2\linewidth]{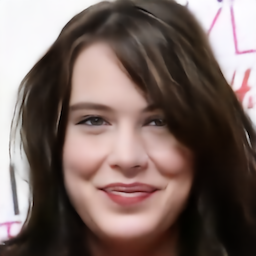}&
    \includegraphics[width=0.2\linewidth]{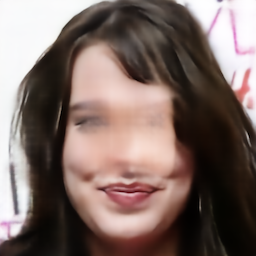}&
    \includegraphics[width=0.2\linewidth]{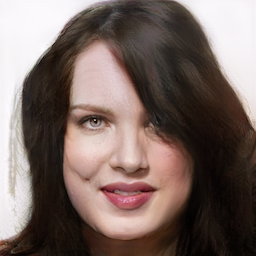}&
    \includegraphics[width=0.2\linewidth]{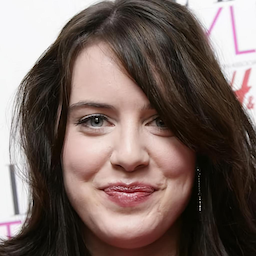}\\[-2.5pt]
    \includegraphics[width=0.2\linewidth]{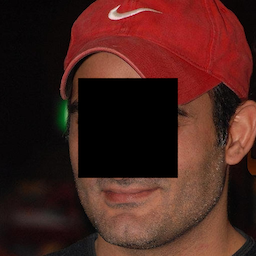}&
    \includegraphics[width=0.2\linewidth]{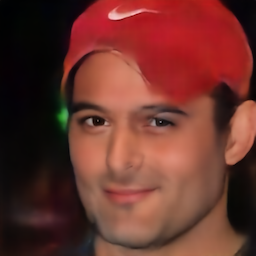}&
    \includegraphics[width=0.2\linewidth]{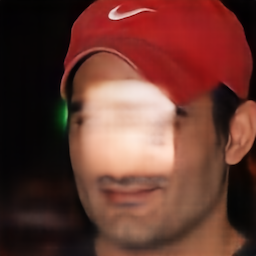}&
    \includegraphics[width=0.2\linewidth]{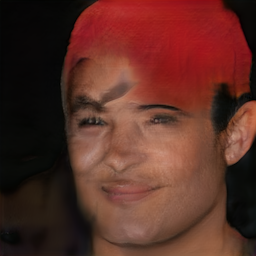}&
    \includegraphics[width=0.2\linewidth]{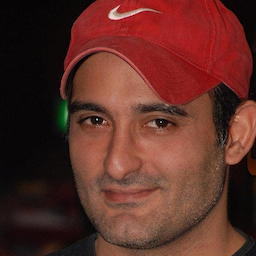}\\[-2.5pt]
    \includegraphics[width=0.2\linewidth]{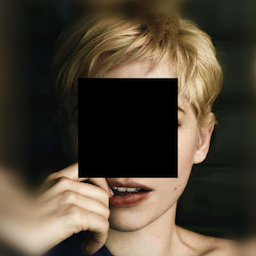}&
    \includegraphics[width=0.2\linewidth]{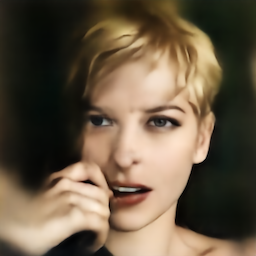}&
    \includegraphics[width=0.2\linewidth]{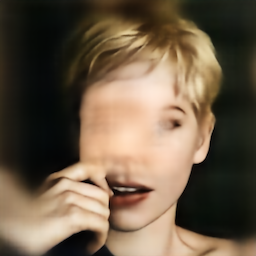}&
    \includegraphics[width=0.2\linewidth]{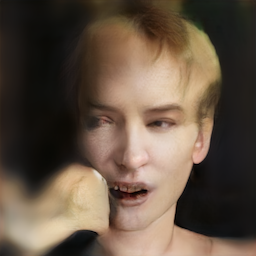}&
    \includegraphics[width=0.2\linewidth]{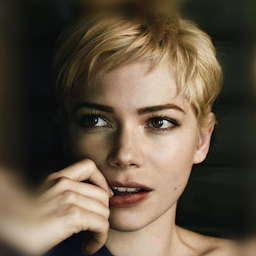}\\[-2.5pt]
    \includegraphics[width=0.2\linewidth]{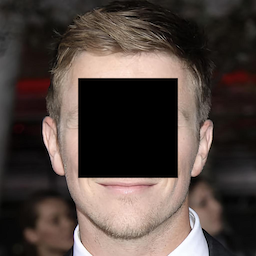}&
    \includegraphics[width=0.2\linewidth]{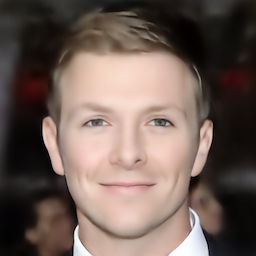}&
    \includegraphics[width=0.2\linewidth]{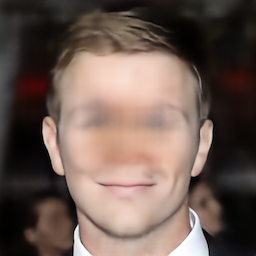}&
    \includegraphics[width=0.2\linewidth]{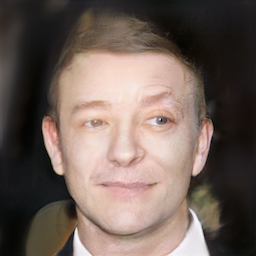}&
    \includegraphics[width=0.2\linewidth]{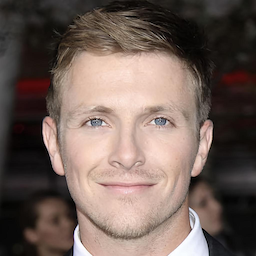}\\[-2.5pt]
    \end{tabular}
\caption{A comparision of optimization over the convolutional manifold (column "OptConv"), the z-space (column "OptZ") and the Progressive GAN \citep{Karras18} latent space (column "PGAN") on the CelebA-HQ dataset \citep{Karras18}.}
\label{fig:ConvMan}
\end{figure}

\subsection{Experiments on CelebA-HQ and the role of the convolutional manifold.} 

For the CelebA-HQ, we have limited comparison of the LCM model to the pretrained  progressive GAN model~\citep{Karras18} published by the authors (this is because proper tuning of the parameters of other baselines would take too much time). On this dataset, LCM uses a latent space of 135k parameters.

Additionally, we use CelebA-HQ to highlight the role of the convolutional manifold structure in the latent space. Recall that the use of the convolutional manifold parameterization is what distinguish the LCM approach from the GLO baseline. The advantage of the new parameterization is highlighted by the experiments described above. One may wonder, if the convolutional manifold constraint is needed at testtime, or if during the restoration process the constraint can be omitted (i.e.\ if \eq{reconstr_baseline} can be used instead of \eq{reconstr} with the generator network $g$ trained with the constraint). Generally, we observed that the use of the constraint at testtime had a minor effect on the CelebA and Bedrooms dataset, but was very pronounced on the CelebA-HQ dataset (where the training set is much smaller and the resolution is much higher). 

In \fig{ConvMan} and \tab{ConvMan}, we provide qualitative and quantitative comparison between the progressive GAN model \citep{Karras18}, the LCM model, and the same LCM model applied without the convolutional manifold constraint for the task of inpainting. The full LCM model with the convolutional manifold performed markedly better than the other two approaches. Progressive GAN severely underfit even the known pixels. This is even despite the fact that the training set of \citep{Karras18} included the validation set (since their model was trained on full CelebA-HQ dataset). Unconstrained LCM overfit the known pixels while providing implausible inpaintings for the unknown. Full LCM model obtained much better balance between fitting the known pixels and inpainting the unknown pixels.

\section{Conclusion}
\label{sect:conclusion}
The results in this work suggest that high-dimensional latent spaces are necessary to get good image reconstructions on desired hold-out sets. Further, it shows that
parametrizing these spaces using ConvNets imposes further structure on them that allow us to produce good image restorations from a wide variety of degradations and at relatively high resolutions. More generally, this method can easily be extended to come up with more interesting parametrizations of the latent space, e.g.\ by interleaving the layers with image-specific and dataset-specific parameters.

The proposed approach has several limitations. First, when trained over very large datasets, the LCM model requires long time to be trained till convergence. For instance, training an LCM on 150k samples of CelebA at \(128\times 128\) resolution takes about 14 GPU-days. Note that the GLO model of the same latent dimensionality takes about 10 GPU-days. On the other hand, the universality of the models means that they only need to be trained once for a certain image type, and can be applied to any degradations after that. The second limiatation is that both LCM and GLO model require storing their latent representations in memory, which for large datasets and large latent spaces may pose a problem. Finally, we observe that even with the large latent dimensionalities that we use here, the models are not able to fit the training data perfectly suffering from such underfitting.



\FloatBarrier

\bibliographystyle{iclr2019_conference}
\bibliography{refs.bib}


\end{document}


\maketitle

\section{Half-Image Inpainting on CelebA}
As another demonstration of the usefulness of our prior, we perform image inpainting with half of the image occluded and with 95\% of its pixels dropped. The results of the restorations are given in \fig{half-inpaint} and \fig{random-inpaint}.
\begin{figure}[h]
    \centering
    \renewcommand{\tabcolsep}{0.5pt}
    \begin{tabular}{ccccccc}
         \small{Hole Image} & \small{LCM} & \small{GLO} & \small{DIP} & \small{WGAN} & \small{AE} &\small{Original Image}\\
         \includegraphics[width=0.14\linewidth]{suppmat_figures/Half_Inpaintings/Hole_Img/Hole_Image8.png}&
         \includegraphics[width=0.14\linewidth]{suppmat_figures/Half_Inpaintings/LCM/Inpainted_Image8.png}&
         \includegraphics[width=0.14\linewidth]{suppmat_figures/Half_Inpaintings/GLO/Inpainted_Image8.png}&
         \includegraphics[width=0.14\linewidth]{suppmat_figures/Half_Inpaintings/DIP/Inpainted_Image8.png}&
         \includegraphics[width=0.14\linewidth]{suppmat_figures/Half_Inpaintings/WGAN-GP/Inpainted_Image8.png}&
         \includegraphics[width=0.14\linewidth]{suppmat_figures/Half_Inpaintings/AE/Inpainted_Image8.png}&
         \includegraphics[width=0.14\linewidth]{suppmat_figures/Half_Inpaintings/Org_Img/Org_image8.png}\\[-2pt]
         \includegraphics[width=0.14\linewidth]{suppmat_figures/Half_Inpaintings/Hole_Img/Hole_Image17.png}&
         \includegraphics[width=0.14\linewidth]{suppmat_figures/Half_Inpaintings/LCM/Inpainted_Image17.png}&
         \includegraphics[width=0.14\linewidth]{suppmat_figures/Half_Inpaintings/GLO/Inpainted_Image17.png}&
         \includegraphics[width=0.14\linewidth]{suppmat_figures/Half_Inpaintings/DIP/Inpainted_Image17.png}&
         \includegraphics[width=0.14\linewidth]{suppmat_figures/Half_Inpaintings/WGAN-GP/Inpainted_Image17.png}&
         \includegraphics[width=0.14\linewidth]{suppmat_figures/Half_Inpaintings/AE/Inpainted_Image17.png}&
         \includegraphics[width=0.14\linewidth]{suppmat_figures/Half_Inpaintings/Org_Img/Org_image17.png}\\[-2pt]
         \includegraphics[width=0.14\linewidth]{suppmat_figures/Half_Inpaintings/Hole_Img/Hole_Image18.png}&
         \includegraphics[width=0.14\linewidth]{suppmat_figures/Half_Inpaintings/LCM/Inpainted_Image18.png}&
         \includegraphics[width=0.14\linewidth]{suppmat_figures/Half_Inpaintings/GLO/Inpainted_Image18.png}&
         \includegraphics[width=0.14\linewidth]{suppmat_figures/Half_Inpaintings/DIP/Inpainted_Image18.png}&
         \includegraphics[width=0.14\linewidth]{suppmat_figures/Half_Inpaintings/WGAN-GP/Inpainted_Image18.png}&
         \includegraphics[width=0.14\linewidth]{suppmat_figures/Half_Inpaintings/AE/Inpainted_Image18.png}&
         \includegraphics[width=0.14\linewidth]{suppmat_figures/Half_Inpaintings/Org_Img/Org_image18.png}\\[-2pt]
         \includegraphics[width=0.14\linewidth]{suppmat_figures/Half_Inpaintings/Hole_Img/Hole_Image22.png}&
         \includegraphics[width=0.14\linewidth]{suppmat_figures/Half_Inpaintings/LCM/Inpainted_Image22.png}&
         \includegraphics[width=0.14\linewidth]{suppmat_figures/Half_Inpaintings/GLO/Inpainted_Image22.png}&
         \includegraphics[width=0.14\linewidth]{suppmat_figures/Half_Inpaintings/DIP/Inpainted_Image22.png}&
         \includegraphics[width=0.14\linewidth]{suppmat_figures/Half_Inpaintings/WGAN-GP/Inpainted_Image22.png}&
         \includegraphics[width=0.14\linewidth]{suppmat_figures/Half_Inpaintings/AE/Inpainted_Image22.png}&
         \includegraphics[width=0.14\linewidth]{suppmat_figures/Half_Inpaintings/Org_Img/Org_image22.png}\\[-2pt]
         \includegraphics[width=0.14\linewidth]{suppmat_figures/Half_Inpaintings/Hole_Img/Hole_Image29.png}&
         \includegraphics[width=0.14\linewidth]{suppmat_figures/Half_Inpaintings/LCM/Inpainted_Image29.png}&
         \includegraphics[width=0.14\linewidth]{suppmat_figures/Half_Inpaintings/GLO/Inpainted_Image29.png}&
         \includegraphics[width=0.14\linewidth]{suppmat_figures/Half_Inpaintings/DIP/Inpainted_Image29.png}&
         \includegraphics[width=0.14\linewidth]{suppmat_figures/Half_Inpaintings/WGAN-GP/Inpainted_Image29.png}&
         \includegraphics[width=0.14\linewidth]{suppmat_figures/Half_Inpaintings/AE/Inpainted_Image29.png}&
         \includegraphics[width=0.14\linewidth]{suppmat_figures/Half_Inpaintings/Org_Img/Org_image29.png}\\[-2pt]
         \includegraphics[width=0.14\linewidth]{suppmat_figures/Half_Inpaintings/Hole_Img/Hole_Image43.png}&
         \includegraphics[width=0.14\linewidth]{suppmat_figures/Half_Inpaintings/LCM/Inpainted_Image43.png}&
         \includegraphics[width=0.14\linewidth]{suppmat_figures/Half_Inpaintings/GLO/Inpainted_Image43.png}&
         \includegraphics[width=0.14\linewidth]{suppmat_figures/Half_Inpaintings/DIP/Inpainted_Image43.png}&
         \includegraphics[width=0.14\linewidth]{suppmat_figures/Half_Inpaintings/WGAN-GP/Inpainted_Image43.png}&
         \includegraphics[width=0.14\linewidth]{suppmat_figures/Half_Inpaintings/AE/Inpainted_Image43.png}&
         \includegraphics[width=0.14\linewidth]{suppmat_figures/Half_Inpaintings/Org_Img/Org_image43.png}
    \end{tabular}
    \caption{Half-image completion task results on the CelebA dataset ($128 \times 128$ resolution)}
    \label{fig:half-inpaint}
\end{figure}
\begin{figure}[H]
    \centering
    \renewcommand{\tabcolsep}{0.5pt}
    \begin{tabular}{cccc}
         \small{Distorted Image} & \small{LCM} & \small{GLO} &\small{Original Image}\\
         \includegraphics[width=0.24\linewidth]{suppmat_figures/RandomMask_Inpainting/Hole_Img/Hole_Image2.png}&
         \includegraphics[width=0.24\linewidth]{suppmat_figures/RandomMask_Inpainting/LCM/Inpainted_Image2.png}&
         \includegraphics[width=0.24\linewidth]{suppmat_figures/RandomMask_Inpainting/GLO/Inpainted_Image2.png}&
         \includegraphics[width=0.24\linewidth]{suppmat_figures/RandomMask_Inpainting/Org_Img/Org_image2.png}\\[-2pt]
         \includegraphics[width=0.24\linewidth]{suppmat_figures/RandomMask_Inpainting/Hole_Img/Hole_Image4.png}&
         \includegraphics[width=0.24\linewidth]{suppmat_figures/RandomMask_Inpainting/LCM/Inpainted_Image4.png}&
         \includegraphics[width=0.24\linewidth]{suppmat_figures/RandomMask_Inpainting/GLO/Inpainted_Image4.png}&
         \includegraphics[width=0.24\linewidth]{suppmat_figures/RandomMask_Inpainting/Org_Img/Org_image4.png}\\[-2pt]
         \includegraphics[width=0.24\linewidth]{suppmat_figures/RandomMask_Inpainting/Hole_Img/Hole_Image5.png}&
         \includegraphics[width=0.24\linewidth]{suppmat_figures/RandomMask_Inpainting/LCM/Inpainted_Image5.png}&
         \includegraphics[width=0.24\linewidth]{suppmat_figures/RandomMask_Inpainting/GLO/Inpainted_Image5.png}&
         \includegraphics[width=0.24\linewidth]{suppmat_figures/RandomMask_Inpainting/Org_Img/Org_image5.png}\\[-2pt]
         \includegraphics[width=0.24\linewidth]{suppmat_figures/RandomMask_Inpainting/Hole_Img/Hole_Image6.png}&
         \includegraphics[width=0.24\linewidth]{suppmat_figures/RandomMask_Inpainting/LCM/Inpainted_Image6.png}&
         \includegraphics[width=0.24\linewidth]{suppmat_figures/RandomMask_Inpainting/GLO/Inpainted_Image6.png}&
         \includegraphics[width=0.24\linewidth]{suppmat_figures/RandomMask_Inpainting/Org_Img/Org_image6.png}\\[-2pt]
         \includegraphics[width=0.24\linewidth]{suppmat_figures/RandomMask_Inpainting/Hole_Img/Hole_Image12.png}&
         \includegraphics[width=0.24\linewidth]{suppmat_figures/RandomMask_Inpainting/LCM/Inpainted_Image12.png}&
         \includegraphics[width=0.24\linewidth]{suppmat_figures/RandomMask_Inpainting/GLO/Inpainted_Image12.png}&
         \includegraphics[width=0.24\linewidth]{suppmat_figures/RandomMask_Inpainting/Org_Img/Org_image12.png}\\[-2pt]
         \includegraphics[width=0.24\linewidth]{suppmat_figures/RandomMask_Inpainting/Hole_Img/Hole_Image10.png}&
         \includegraphics[width=0.24\linewidth]{suppmat_figures/RandomMask_Inpainting/LCM/Inpainted_Image10.png}&
         \includegraphics[width=0.24\linewidth]{suppmat_figures/RandomMask_Inpainting/GLO/Inpainted_Image10.png}&
         \includegraphics[width=0.24\linewidth]{suppmat_figures/RandomMask_Inpainting/Org_Img/Org_image10.png}\\[-2pt]
    \end{tabular}
    \caption{Image inpainting on CelebA with \(95\%\) of randomly chosen pixels missing.}
    \label{fig:random-inpaint}
\end{figure}

\section{GLO-latent Inpaintings}
In this section, we show the results of using extremely high dimensional vectors as input to the GLO model (thus, reproducing the original system of Bojanowski et al.). The results are in \fig{glo_recons} and Table \ref{table:Recons-lossGLO}. Here columns denoted by $\mathrm{xxxx\,(vec)}$ correspond to the GLO models with usual vectorial inputs of high-dimension. The column $\mathrm{24k\,(Conv)}$ is the GLO baseline model, where the generator network has the same architecture as in LCM, but the convolutional space is parameterized by a set of maps with $\mathrm{24k}$ parameters. We see that the vector based GLO model, despite being trained on latent vector with relatively high dimensionality, underfit the images to be inpainted. Unfortunately, due to memory constraints we could not fit a $\mathrm{24k}$-dimensional vector-based GLO model for a more direct comparison.
\begin{table}[h]
 \centering
 \begin{tabular}{@{}ccc@{}}
 \toprule
 \textbf{Latent Space Type} & \textbf{Latent Space Dimension} & \textbf{Reconstruction Loss} \\ \midrule
 3D Map            & 24576                  &    0.0113                          \\ 
 Vector                 & 8192                  & 0.0144                              \\ 
 Vector                & 4096                  & 0.0171                             \\ 
 Vector                 & 2048                  & 0.0202                              \\ 
 \bottomrule\\
 \end{tabular}
 \caption{Reconstruction losses of different GLO variants.}
 \label{table:Recons-lossGLO}
\end{table}
\begin{figure}[h!]
    \centering
    \renewcommand{\tabcolsep}{0.5pt}
    \begin{tabular}{cccccc}
         \small{Hole Image} & \small{24k (Conv)} & \small{8192 (vec)} & \small{4096 (vec)} & \small{2048 (vec)} & \small{Original Image}\\
         \includegraphics[width=0.15\linewidth]{suppmat_figures/GLO_Latent/Hole_Img/Hole_Image8.png}&
         \includegraphics[width=0.15\linewidth]{suppmat_figures/GLO_Latent/GLO_Conv/Inpainted_Image8.png}&
         \includegraphics[width=0.15\linewidth]{suppmat_figures/GLO_Latent/GLO_8192/Inpainted_Image8.png}&
         \includegraphics[width=0.15\linewidth]{suppmat_figures/GLO_Latent/GLO_4096/Inpainted_Image8.png}&
         \includegraphics[width=0.15\linewidth]{suppmat_figures/GLO_Latent/GLO_2048/Inpainted_Image8.png}&
         \includegraphics[width=0.15\linewidth]{suppmat_figures/GLO_Latent/Org_Img/Org_image8.png}\\[-2pt]
         \includegraphics[width=0.15\linewidth]{suppmat_figures/GLO_Latent/Hole_Img/Hole_Image10.png}&
         \includegraphics[width=0.15\linewidth]{suppmat_figures/GLO_Latent/GLO_Conv/Inpainted_Image10.png}&
         \includegraphics[width=0.15\linewidth]{suppmat_figures/GLO_Latent/GLO_8192/Inpainted_Image10.png}&
         \includegraphics[width=0.15\linewidth]{suppmat_figures/GLO_Latent/GLO_4096/Inpainted_Image10.png}&
         \includegraphics[width=0.15\linewidth]{suppmat_figures/GLO_Latent/GLO_2048/Inpainted_Image10.png}&
         \includegraphics[width=0.15\linewidth]{suppmat_figures/GLO_Latent/Org_Img/Org_image10.png}\\[-2pt]
         \includegraphics[width=0.15\linewidth]{suppmat_figures/GLO_Latent/Hole_Img/Hole_Image26.png}&
         \includegraphics[width=0.15\linewidth]{suppmat_figures/GLO_Latent/GLO_Conv/Inpainted_Image26.png}&
         \includegraphics[width=0.15\linewidth]{suppmat_figures/GLO_Latent/GLO_8192/Inpainted_Image26.png}&
         \includegraphics[width=0.15\linewidth]{suppmat_figures/GLO_Latent/GLO_4096/Inpainted_Image26.png}&
         \includegraphics[width=0.15\linewidth]{suppmat_figures/GLO_Latent/GLO_2048/Inpainted_Image26.png}&
         \includegraphics[width=0.15\linewidth]{suppmat_figures/GLO_Latent/Org_Img/Org_image26.png}\\[-2pt]
         \includegraphics[width=0.15\linewidth]{suppmat_figures/GLO_Latent/Hole_Img/Hole_Image29.png}&
         \includegraphics[width=0.15\linewidth]{suppmat_figures/GLO_Latent/GLO_Conv/Inpainted_Image29.png}&
         \includegraphics[width=0.15\linewidth]{suppmat_figures/GLO_Latent/GLO_8192/Inpainted_Image29.png}&
         \includegraphics[width=0.15\linewidth]{suppmat_figures/GLO_Latent/GLO_4096/Inpainted_Image29.png}&
         \includegraphics[width=0.15\linewidth]{suppmat_figures/GLO_Latent/GLO_2048/Inpainted_Image29.png}&
         \includegraphics[width=0.15\linewidth]{suppmat_figures/GLO_Latent/Org_Img/Org_image29.png}\\[-2pt]
         \includegraphics[width=0.15\linewidth]{suppmat_figures/GLO_Latent/Hole_Img/Hole_Image30.png}&
         \includegraphics[width=0.15\linewidth]{suppmat_figures/GLO_Latent/GLO_Conv/Inpainted_Image30.png}&
         \includegraphics[width=0.15\linewidth]{suppmat_figures/GLO_Latent/GLO_8192/Inpainted_Image30.png}&
         \includegraphics[width=0.15\linewidth]{suppmat_figures/GLO_Latent/GLO_4096/Inpainted_Image30.png}&
         \includegraphics[width=0.15\linewidth]{suppmat_figures/GLO_Latent/GLO_2048/Inpainted_Image30.png}&
         \includegraphics[width=0.15\linewidth]{suppmat_figures/GLO_Latent/Org_Img/Org_image30.png}\\[-2pt]
         \includegraphics[width=0.15\linewidth]{suppmat_figures/GLO_Latent/Hole_Img/Hole_Image44.png}&
         \includegraphics[width=0.15\linewidth]{suppmat_figures/GLO_Latent/GLO_Conv/Inpainted_Image44.png}&
         \includegraphics[width=0.15\linewidth]{suppmat_figures/GLO_Latent/GLO_8192/Inpainted_Image44.png}&
         \includegraphics[width=0.15\linewidth]{suppmat_figures/GLO_Latent/GLO_4096/Inpainted_Image44.png}&
         \includegraphics[width=0.15\linewidth]{suppmat_figures/GLO_Latent/GLO_2048/Inpainted_Image44.png}&
         \includegraphics[width=0.15\linewidth]{suppmat_figures/GLO_Latent/Org_Img/Org_image44.png}\\[-2pt]
    \end{tabular}
    \caption{Image inpainting using GLO models with latent spaces of different dimension and structure.}
    \label{fig:glo_recons}
\end{figure}

\section{WGAN-Reconstructions}
Given below are the reconstruction losses for different values of L2 penalty on the latent vector on the CelebA dataset using a WGAN-GP. We observe that increasing the L2-penalty on the latent variable increases the reconstruction losses without providing any major improvement in the quality of the fits. Thus, we decided not to regularize the latent variables while performing image restoration on various tasks. 

To the best of our knowledge, apart from the progressive GAN, no other GAN-based system reported good performance on 128x128 face images that were not tightly cropped to remove most of background (which simplifies modeling considerably).

\begin{table}[h]
 \centering
 \begin{tabular}{@{}cc@{}}
 \toprule
 \textbf{L2-penalty Weight} & \textbf{Reconstruction Loss} \\ \midrule
 1            & 0.53                  \\ 
 \(1e^{-3}\)                 & 0.3178       \\ 
 \(1e^{-5}\)                & 0.2176        \\
 \(0\)                & 0.1893       \\
 \bottomrule\\
 \end{tabular}
 \caption{Reconstruction losses with different weight penalties using the WGAN-GP.}
 \label{table:Recons-lossWGAN}
\end{table}

\begin{figure}[h]
    \centering
    \renewcommand{\tabcolsep}{0.5pt}
    \begin{tabular}{ccccc}
         \small{Original Image} & \small{0} & \small{1e-5} & \small{1e-3} & \small{1}\\
         \includegraphics[width=0.2\linewidth]{suppmat_figures/WGAN_Fits/Org_Img/Org_image0.png}&
         \includegraphics[width=0.2\linewidth]{suppmat_figures/WGAN_Fits/0/Inpainted_Image0.png}&
         \includegraphics[width=0.2\linewidth]{suppmat_figures/WGAN_Fits/1e-5/Inpainted_Image0.png}&
         \includegraphics[width=0.2\linewidth]{suppmat_figures/WGAN_Fits/1e-3/Inpainted_Image0.png}&
         \includegraphics[width=0.2\linewidth]{suppmat_figures/WGAN_Fits/1/Inpainted_Image0.png}\\[-2pt]
         \includegraphics[width=0.2\linewidth]{suppmat_figures/WGAN_Fits/Org_Img/Org_image8.png}&
         \includegraphics[width=0.2\linewidth]{suppmat_figures/WGAN_Fits/0/Inpainted_Image8.png}&
         \includegraphics[width=0.2\linewidth]{suppmat_figures/WGAN_Fits/1e-5/Inpainted_Image8.png}&
         \includegraphics[width=0.2\linewidth]{suppmat_figures/WGAN_Fits/1e-3/Inpainted_Image8.png}&
         \includegraphics[width=0.2\linewidth]{suppmat_figures/WGAN_Fits/1/Inpainted_Image8.png}\\[-2pt]
         \includegraphics[width=0.2\linewidth]{suppmat_figures/WGAN_Fits/Org_Img/Org_image29.png}&
         \includegraphics[width=0.2\linewidth]{suppmat_figures/WGAN_Fits/0/Inpainted_Image29.png}&
         \includegraphics[width=0.2\linewidth]{suppmat_figures/WGAN_Fits/1e-5/Inpainted_Image29.png}&
         \includegraphics[width=0.2\linewidth]{suppmat_figures/WGAN_Fits/1e-3/Inpainted_Image29.png}&
         \includegraphics[width=0.2\linewidth]{suppmat_figures/WGAN_Fits/1/Inpainted_Image29.png}\\[-2pt]
         \includegraphics[width=0.2\linewidth]{suppmat_figures/WGAN_Fits/Org_Img/Org_image39.png}&
         \includegraphics[width=0.2\linewidth]{suppmat_figures/WGAN_Fits/0/Inpainted_Image39.png}&
         \includegraphics[width=0.2\linewidth]{suppmat_figures/WGAN_Fits/1e-5/Inpainted_Image39.png}&
         \includegraphics[width=0.2\linewidth]{suppmat_figures/WGAN_Fits/1e-3/Inpainted_Image39.png}&
         \includegraphics[width=0.2\linewidth]{suppmat_figures/WGAN_Fits/1/Inpainted_Image39.png}\\[-2pt]
    \end{tabular}
    \caption{Image reconstruction using the WGAN-GP with different L2 penalties.}
    \label{fig:wgan_recons}
\end{figure}
\section{Network Details}

We used the following networks in our model:

\begin{itemize}
    \item{Generator Network \(g_{\theta}\)}: The generator network \(g_{\theta}\) has an hourglass architecture in all three datasets. In CelebA the map size varies as follows: \(32\times{}32 \rightarrow 4\times4 \rightarrow 128\times128\). In Bedrooms the map size varies as: \(64\times{}64 \rightarrow 4\times4 \rightarrow 256\times256\). In CelebAHQ the map size varies as \(256\times{}256 \rightarrow 32\times32 \rightarrow 1024\times1024\). All the generator networks contain two skip connections within them.
    \item{Latent Network \(f_{\phi_{i}}\)}: The latent network used in CelebA128 consists of 4 convolutional layers with no padding. The latent network used in Bedrooms and CelebAHQ consists of 5 convolutional layers with no padding.
\end{itemize}

\section{More Results on CelebAHQ}
Here we provide the comparison between LCM and GLO for CelebA-HQ. \begin{figure}[H]
    \centering
    \renewcommand{\tabcolsep}{0.5pt}
    \begin{tabular}{cccc}
         \small{Distorted Image} & \small{LCM} & \small{GLO} & \small{Original Image}\\
         \includegraphics[width=0.2\linewidth]{suppmat_figures/CelebHQ/Hole_Img/Hole_Image6.png}&
         \includegraphics[width=0.2\linewidth]{suppmat_figures/CelebHQ/LCM/Inpainted_Image6.png}&
         \includegraphics[width=0.2\linewidth]{suppmat_figures/CelebHQ/GLO/Inpainted_Image6.png}&
         \includegraphics[width=0.2\linewidth]{suppmat_figures/CelebHQ/Org_Img/Org_image6.png}\\[-2pt]
         \includegraphics[width=0.2\linewidth]{suppmat_figures/CelebHQ/Hole_Img/Hole_Image17.png}&
         \includegraphics[width=0.2\linewidth]{suppmat_figures/CelebHQ/LCM/Inpainted_Image17.png}&
         \includegraphics[width=0.2\linewidth]{suppmat_figures/CelebHQ/GLO/Inpainted_Image17.png}&
         \includegraphics[width=0.2\linewidth]{suppmat_figures/CelebHQ/Org_Img/Org_image17.png}\\[-2pt]
         \includegraphics[width=0.2\linewidth]{suppmat_figures/CelebHQ/Hole_Img/Hole_Image20.png}&
         \includegraphics[width=0.2\linewidth]{suppmat_figures/CelebHQ/LCM/Inpainted_Image20.png}&
         \includegraphics[width=0.2\linewidth]{suppmat_figures/CelebHQ/GLO/Inpainted_Image20.png}&
         \includegraphics[width=0.2\linewidth]{suppmat_figures/CelebHQ/Org_Img/Org_image20.png}\\[-2pt]
         \includegraphics[width=0.2\linewidth]{suppmat_figures/CelebHQ/Hole_Img/Hole_Image26.png}&
         \includegraphics[width=0.2\linewidth]{suppmat_figures/CelebHQ/LCM/Inpainted_Image26.png}&
         \includegraphics[width=0.2\linewidth]{suppmat_figures/CelebHQ/GLO/Inpainted_Image26.png}&
         \includegraphics[width=0.2\linewidth]{suppmat_figures/CelebHQ/Org_Img/Org_image26.png}\\[-2pt]
         \includegraphics[width=0.2\linewidth]{suppmat_figures/CelebHQ/Hole_Img/Hole_Image27.png}&
         \includegraphics[width=0.2\linewidth]{suppmat_figures/CelebHQ/LCM/Inpainted_Image27.png}&
         \includegraphics[width=0.2\linewidth]{suppmat_figures/CelebHQ/GLO/Inpainted_Image27.png}&
         \includegraphics[width=0.2\linewidth]{suppmat_figures/CelebHQ/Org_Img/Org_image27.png}\\[-2pt]
         \includegraphics[width=0.2\linewidth]{suppmat_figures/CelebHQ/Hole_Img/Hole_Image33.png}&
         \includegraphics[width=0.2\linewidth]{suppmat_figures/CelebHQ/LCM/Inpainted_Image33.png}&
         \includegraphics[width=0.2\linewidth]{suppmat_figures/CelebHQ/GLO/Inpainted_Image33.png}&
         \includegraphics[width=0.2\linewidth]{suppmat_figures/CelebHQ/Org_Img/Org_image33.png}\\[-2pt]
         \includegraphics[width=0.2\linewidth]{suppmat_figures/CelebHQ/Hole_Img/Hole_Image40.png}&
         \includegraphics[width=0.2\linewidth]{suppmat_figures/CelebHQ/LCM/Inpainted_Image40.png}&
         \includegraphics[width=0.2\linewidth]{suppmat_figures/CelebHQ/GLO/Inpainted_Image40.png}&
         \includegraphics[width=0.2\linewidth]{suppmat_figures/CelebHQ/Org_Img/Org_image40.png}\\[-2pt]
         \includegraphics[width=0.2\linewidth]{suppmat_figures/CelebHQ/Hole_Img/Hole_Image43.png}&
         \includegraphics[width=0.2\linewidth]{suppmat_figures/CelebHQ/LCM/Inpainted_Image43.png}&
         \includegraphics[width=0.2\linewidth]{suppmat_figures/CelebHQ/GLO/Inpainted_Image43.png}&
         \includegraphics[width=0.2\linewidth]{suppmat_figures/CelebHQ/Org_Img/Org_image43.png}\\[-2pt]
    \end{tabular}
\end{figure}

\section{More Results on Bedrooms}

\begin{figure}[H]
    \centering
    \renewcommand{\tabcolsep}{0.5pt}
    \begin{tabular}{cccccc}
         \small{Distorted Image} & \small{LCM} & \small{GLO} & \small{DIP} & \small{PGAN} &  \small{Original Image}\\
         \includegraphics[width=0.16\linewidth]{suppmat_figures/Bedrooms/Inpainting/Hole_Img/Hole_Image16.png}&
         \includegraphics[width=0.16\linewidth]{suppmat_figures/Bedrooms/Inpainting/LCM/Inpainted_Image16.png}&
         \includegraphics[width=0.16\linewidth]{suppmat_figures/Bedrooms/Inpainting/GLO/Inpainted_Image16.png}&
         \includegraphics[width=0.16\linewidth]{suppmat_figures/Bedrooms/Inpainting/DIP/Inpainted_Image16.png}&
         \includegraphics[width=0.16\linewidth]{suppmat_figures/Bedrooms/Inpainting/PGAN/Inpainted_Image16.png}&
         \includegraphics[width=0.16\linewidth]{suppmat_figures/Bedrooms/Inpainting/Org_Img/Org_image16.png}\\[-2pt]
         \includegraphics[width=0.16\linewidth]{suppmat_figures/Bedrooms/Inpainting/Hole_Img/Hole_Image17.png}&
         \includegraphics[width=0.16\linewidth]{suppmat_figures/Bedrooms/Inpainting/LCM/Inpainted_Image17.png}&
         \includegraphics[width=0.16\linewidth]{suppmat_figures/Bedrooms/Inpainting/GLO/Inpainted_Image17.png}&
         \includegraphics[width=0.16\linewidth]{suppmat_figures/Bedrooms/Inpainting/DIP/Inpainted_Image17.png}&
         \includegraphics[width=0.16\linewidth]{suppmat_figures/Bedrooms/Inpainting/PGAN/Inpainted_Image17.png}&
         \includegraphics[width=0.16\linewidth]{suppmat_figures/Bedrooms/Inpainting/Org_Img/Org_image17.png}\\[-2pt]
         \includegraphics[width=0.16\linewidth]{suppmat_figures/Bedrooms/Inpainting/Hole_Img/Hole_Image23.png}&
         \includegraphics[width=0.16\linewidth]{suppmat_figures/Bedrooms/Inpainting/LCM/Inpainted_Image23.png}&
         \includegraphics[width=0.16\linewidth]{suppmat_figures/Bedrooms/Inpainting/GLO/Inpainted_Image23.png}&
         \includegraphics[width=0.16\linewidth]{suppmat_figures/Bedrooms/Inpainting/DIP/Inpainted_Image23.png}&
         \includegraphics[width=0.16\linewidth]{suppmat_figures/Bedrooms/Inpainting/PGAN/Inpainted_Image23.png}&
         \includegraphics[width=0.16\linewidth]{suppmat_figures/Bedrooms/Inpainting/Org_Img/Org_image23.png}\\[-2pt]
         \includegraphics[width=0.16\linewidth]{suppmat_figures/Bedrooms/Inpainting/Hole_Img/Hole_Image24.png}&
         \includegraphics[width=0.16\linewidth]{suppmat_figures/Bedrooms/Inpainting/LCM/Inpainted_Image24.png}&
         \includegraphics[width=0.16\linewidth]{suppmat_figures/Bedrooms/Inpainting/GLO/Inpainted_Image24.png}&
         \includegraphics[width=0.16\linewidth]{suppmat_figures/Bedrooms/Inpainting/DIP/Inpainted_Image24.png}&
         \includegraphics[width=0.16\linewidth]{suppmat_figures/Bedrooms/Inpainting/PGAN/Inpainted_Image24.png}&
         \includegraphics[width=0.16\linewidth]{suppmat_figures/Bedrooms/Inpainting/Org_Img/Org_image24.png}\\[-2pt]
         \includegraphics[width=0.16\linewidth]{suppmat_figures/Bedrooms/Inpainting/Hole_Img/Hole_Image33.png}&
         \includegraphics[width=0.16\linewidth]{suppmat_figures/Bedrooms/Inpainting/LCM/Inpainted_Image33.png}&
         \includegraphics[width=0.16\linewidth]{suppmat_figures/Bedrooms/Inpainting/GLO/Inpainted_Image33.png}&
         \includegraphics[width=0.16\linewidth]{suppmat_figures/Bedrooms/Inpainting/DIP/Inpainted_Image33.png}&
         \includegraphics[width=0.16\linewidth]{suppmat_figures/Bedrooms/Inpainting/PGAN/Inpainted_Image33.png}&
         \includegraphics[width=0.16\linewidth]{suppmat_figures/Bedrooms/Inpainting/Org_Img/Org_image33.png}\\[-2pt]
    \end{tabular}
\end{figure}

\begin{figure}[H]
    \centering
    \renewcommand{\tabcolsep}{0.5pt}
    \begin{tabular}{cccccc}
         \small{Distorted Image} & \small{LCM} & \small{GLO} & \small{DIP} & \small{PGAN} &  \small{Original Image}\\
         \includegraphics[width=0.16\linewidth]{suppmat_figures/Bedrooms/SuperRes/Down_Img/Down_Image16.png}&
         \includegraphics[width=0.16\linewidth]{suppmat_figures/Bedrooms/SuperRes/LCM/Up_Image16.png}&
         \includegraphics[width=0.16\linewidth]{suppmat_figures/Bedrooms/SuperRes/GLO/Up_Image16.png}&
         \includegraphics[width=0.16\linewidth]{suppmat_figures/Bedrooms/SuperRes/DIP/Up_Image16.png}&
         \includegraphics[width=0.16\linewidth]{suppmat_figures/Bedrooms/SuperRes/PGAN/Up_Image16.png}&
         \includegraphics[width=0.16\linewidth]{suppmat_figures/Bedrooms/SuperRes/Org_Img/Org_image16.png}\\[-2pt]
         \includegraphics[width=0.16\linewidth]{suppmat_figures/Bedrooms/SuperRes/Down_Img/Down_Image17.png}&
         \includegraphics[width=0.16\linewidth]{suppmat_figures/Bedrooms/SuperRes/LCM/Up_Image17.png}&
         \includegraphics[width=0.16\linewidth]{suppmat_figures/Bedrooms/SuperRes/GLO/Up_Image17.png}&
         \includegraphics[width=0.16\linewidth]{suppmat_figures/Bedrooms/SuperRes/DIP/Up_Image17.png}&
         \includegraphics[width=0.16\linewidth]{suppmat_figures/Bedrooms/SuperRes/PGAN/Up_Image17.png}&
         \includegraphics[width=0.16\linewidth]{suppmat_figures/Bedrooms/SuperRes/Org_Img/Org_image17.png}\\[-2pt]
         \includegraphics[width=0.16\linewidth]{suppmat_figures/Bedrooms/SuperRes/Down_Img/Down_Image23.png}&
         \includegraphics[width=0.16\linewidth]{suppmat_figures/Bedrooms/SuperRes/LCM/Up_Image23.png}&
         \includegraphics[width=0.16\linewidth]{suppmat_figures/Bedrooms/SuperRes/GLO/Up_Image23.png}&
         \includegraphics[width=0.16\linewidth]{suppmat_figures/Bedrooms/SuperRes/DIP/Up_Image23.png}&
         \includegraphics[width=0.16\linewidth]{suppmat_figures/Bedrooms/SuperRes/PGAN/Up_Image23.png}&
         \includegraphics[width=0.16\linewidth]{suppmat_figures/Bedrooms/SuperRes/Org_Img/Org_image23.png}\\[-2pt]
         \includegraphics[width=0.16\linewidth]{suppmat_figures/Bedrooms/SuperRes/Down_Img/Down_Image24.png}&
         \includegraphics[width=0.16\linewidth]{suppmat_figures/Bedrooms/SuperRes/LCM/Up_Image24.png}&
         \includegraphics[width=0.16\linewidth]{suppmat_figures/Bedrooms/SuperRes/GLO/Up_Image24.png}&
         \includegraphics[width=0.16\linewidth]{suppmat_figures/Bedrooms/SuperRes/DIP/Up_Image24.png}&
         \includegraphics[width=0.16\linewidth]{suppmat_figures/Bedrooms/SuperRes/PGAN/Up_Image24.png}&
         \includegraphics[width=0.16\linewidth]{suppmat_figures/Bedrooms/SuperRes/Org_Img/Org_image24.png}\\[-2pt]
         \includegraphics[width=0.16\linewidth]{suppmat_figures/Bedrooms/SuperRes/Down_Img/Down_Image33.png}&
         \includegraphics[width=0.16\linewidth]{suppmat_figures/Bedrooms/SuperRes/LCM/Up_Image33.png}&
         \includegraphics[width=0.16\linewidth]{suppmat_figures/Bedrooms/SuperRes/GLO/Up_Image33.png}&
         \includegraphics[width=0.16\linewidth]{suppmat_figures/Bedrooms/SuperRes/DIP/Up_Image33.png}&
         \includegraphics[width=0.16\linewidth]{suppmat_figures/Bedrooms/SuperRes/PGAN/Up_Image33.png}&
         \includegraphics[width=0.16\linewidth]{suppmat_figures/Bedrooms/SuperRes/Org_Img/Org_image33.png}
    \end{tabular}
\end{figure}